\documentclass[journal]{IEEEtran}

\usepackage[utf8]{inputenc} 
\usepackage[T1]{fontenc}    
\usepackage{graphicx} \graphicspath{ {figures/} }
\usepackage{amsmath,amssymb}
\usepackage{times}
\usepackage{graphicx}
\usepackage{algorithmic}
\usepackage[ruled]{algorithm2e}
\usepackage{acronym}
\usepackage{enumitem}
\usepackage[breaklinks,colorlinks,linkcolor=red,citecolor=blue]{hyperref}
\usepackage{balance}
\usepackage{xspace,setspace}
\usepackage{setspace}
\usepackage[skip=3pt,font=small]{subcaption}
\usepackage[skip=3pt,font=small]{caption}
\usepackage[dvipsnames,svgnames,x11names]{xcolor}
\usepackage[capitalise,nameinlink]{cleveref}
\usepackage{booktabs,tabularx,colortbl,multirow,multicol,array,makecell}
\usepackage{cuted}
\usepackage{capt-of}
\usepackage[misc]{ifsym}
\usepackage{pifont}
\usepackage{cite}
\usepackage{listings}
\usepackage{anyfontsize}
\usepackage{tikz}
\usepackage{tcolorbox}

\newcommand{\framedtext}[1]{%
\par\vspace{1mm} 
\noindent\fbox{%
    \parbox{\dimexpr\linewidth-2\fboxsep-2\fboxrule}{#1}%
}%
\par\vspace{2mm} 
}

\makeatletter
\DeclareRobustCommand\onedot{\futurelet\@let@token\@onedot}
\def\@onedot{\ifx\@let@token.\else.\null\fi\xspace}
\def\eg{\emph{e.g}\onedot} 

\def\ie{\emph{i.e}\onedot}

\makeatother

\makeatletter
\def\BState{\State\hskip-\ALG@thistlm}
\makeatother

\makeatletter
\renewcommand{\paragraph}{%
  \@startsection{paragraph}{4}%
  {\z@}{0ex \@plus 0ex \@minus 0ex}{-1em}%
  {\hskip\parindent\normalfont\normalsize\bfseries}%
}
\makeatother

\makeatletter
\AtBeginDocument
 {
   \def\ltx@label#1{\cref@label{#1}}
   \def\label@in@display@noarg#1{\cref@old@label@in@display{#1}}
\def\label@in@mmeasure@noarg#1{%
    \begingroup%
      \measuring@false%
      \cref@old@label@in@display{#1}
    \endgroup}%
 } %
\makeatother

\crefname{algorithm}{Alg.}{Algs.}
\Crefname{algocf}{Algorithm}{Algorithms}
\crefname{section}{Sec.}{Secs.}
\Crefname{section}{Section}{Sections}
\crefname{table}{Tab.}{Tabs.}
\Crefname{table}{Table}{Tables}
\crefname{figure}{Fig.}{Figs.}
\crefname{equation}{Eq.}{Eqs.}
\Crefname{equation}{Equation}{Equations}
\Crefname{figure}{Figure}{Figures}
\crefname{appendix}{Appendix}{Appendices}
\Crefname{appendix}{Appendix}{Appendices}

\definecolor{gblue}{HTML}{4285F4}
\definecolor{gred}{HTML}{DB4437}
\definecolor{ggreen}{HTML}{0F9D58}

\definecolor{mygray}{gray}{.92}

\acrodef{dof}[DoF]{Degree of Freedom}
\acrodef{vkc}[VKC]{Virtual Kinematic Chain}
\acrodef{tamp}[TAMP]{Task and Motion Planning}
\acrodef{pddl}[PDDL]{Planning Domain Definition Language}
\acrodef{rddl}[RDDL]{Relational Dynamic Influence Diagram Language}
\acrodef{htn}[HTN]{Hierarchical Task Network}
\acrodef{rrt}[RRT]{Rapidly-exploring Random Tree}
\acrodef{ompl}[OMPL]{Open Motion Planning Library}
\acrodef{iws}[IWS]{Iterated Width Search}
\acrodef{bfs}[BFS]{Breadth First Search}
\acrodef{ai}[AI]{Artificial Intelligence}
\acrodef{spt}[SPT]{Scene Parse Tree}
\acrodef{com}[CoM]{Center of Mass}
\acrodef{mcmc}[MCMC]{Markov Chain Monte Carlo}
\acrodef{ged}[GED]{Graph Editing Distance}
\acrodef{ga}[GA]{Genetic Algorithm}
\acrodef{sgd}[SGD]{Stochastic Gradient Descent}
\acrodef{llm}[LLM]{Large Language Model}
\acrodef{cg}[$cg$]{contact graph}
\acrodef{cga}[$cg^\oplus$]{augmented contact graph}
\acrodef{sdf}[SDF]{Signed Distance Field}
\acrodef{minlp}[MINLP]{mixed-integer nonlinear program}
\acrodef{cmaes}[CMA-ES]{Covariance Matrix Adaptation Evolution Strategy}
\acrodef{vla}[VLA]{Vision-Language-Action}
\acrodef{vqvae}[VQ-VAE]{Vector Quantized Variational Autoencoder}
\acrodef{vae}[VAE]{Variational Autoencoder}
\acrodef{vbts}[VBTS]{Vision-based Tactile Sensor}
\acrodef{act}[ACT]{Action Chunking with Transformers}

\newcommand{\modelname}{TaF-VLA\xspace}

\newcommand{\authorblockN}[1]{#1}

\usepackage{booktabs}
\usepackage{colortbl}
\usepackage{xcolor}

\definecolor{lightred}{RGB}{255,235,235}
\definecolor{lightred2}{RGB}{255,245,245}
\definecolor{lightblue}{RGB}{235,245,255}
\definecolor{lightblue2}{RGB}{245,250,255}
\definecolor{actbg}{RGB}{255, 240, 240} 
\definecolor{dpbg}{RGB}{240, 255, 240}  
\definecolor{vlabg}{RGB}{240, 245, 255} 

\title{Tactile-Force Alignment in Vision-Language-Action Models for Force-aware Manipulation}

\author{
    Yuzhe Huang$^{1,3\star}$,\quad
    Pei Lin$^{2,3\star}$,\quad
    Wanlin Li$^{3\star}$,\quad
    Daohan Li$^{3}$,\quad
    Jiajun Li$^{3,4}$,\quad
    Jiaming Jiang$^{2,3}$,\\
    Chenxi Xiao$^{2\dagger}$,\quad
    Ziyuan Jiao$^{1,3\dagger}$

    \authorblockN{
    $^1$Beihang University\quad
    $^2$ShanghaiTech University\\
    $^3$Beijing Institute for General Artificial Intelligence\quad
    $^4$The University of Hong Kong
    }

    \authorblockN{
    $^\star$equal contributors\quad 
    $^\dagger$corresponding authors
    }
    
    \href{https://peilin-666.github.io/projects/TaF_VLA/}{https://peilin-666.github.io/projects/TaF-VLA/}
    }

\begin{document}

\maketitle

\begin{figure*}
    \centering
    \includegraphics[width=1\textwidth]{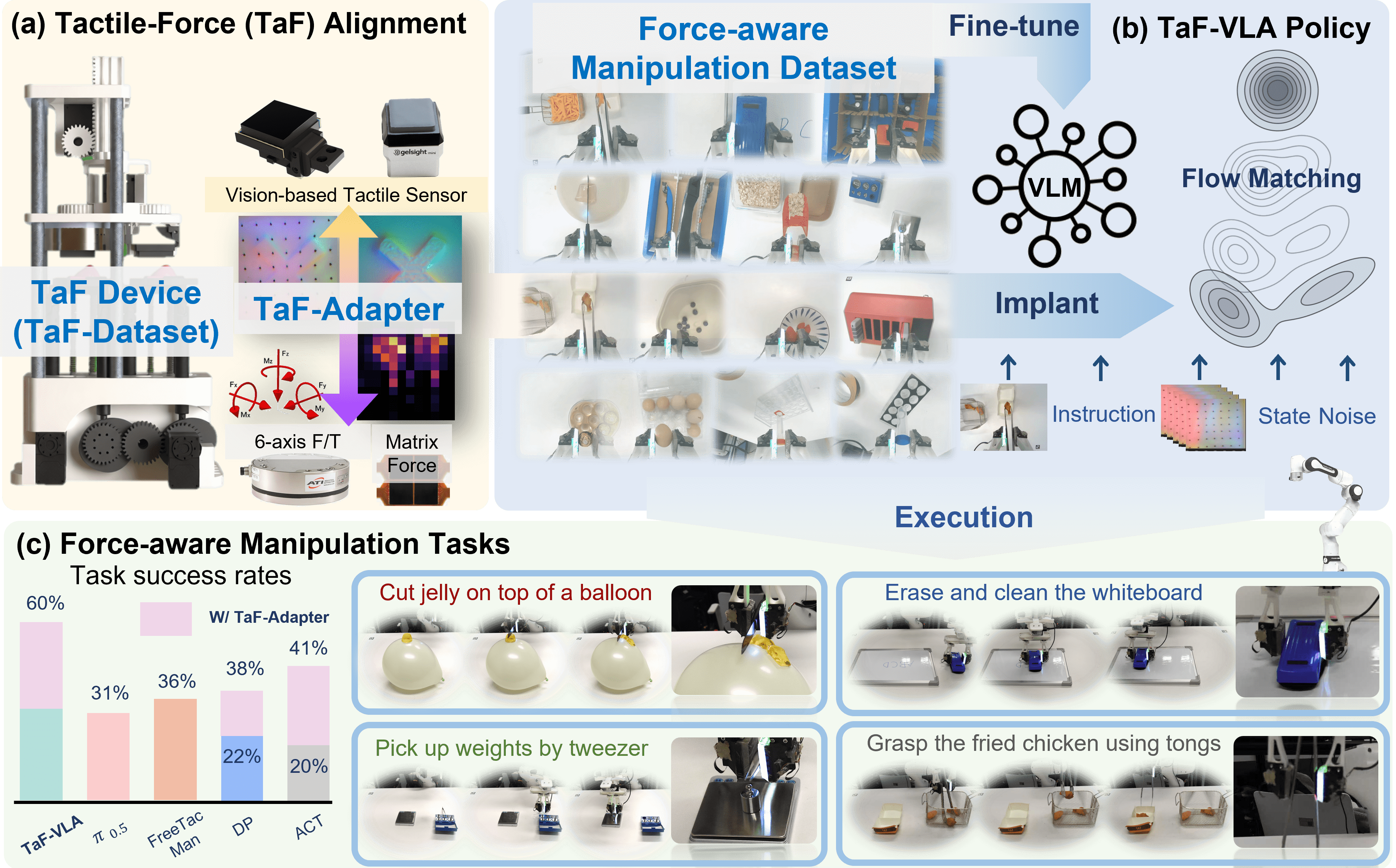}
    \caption{\textbf{From Data to Policy: The \modelname Pipeline.} 
     To address the ``force-blindness'' of current \ac{vla} models, we propose a paradigm shift from tactile-vision to tactile-force alignment, realized through three stages:
     (a) We deploy an automated data acquisition system (TaF-Device) to construct the TaF-Dataset, a large-scale collection of synchronized visuotactile images, 6-axis force/torque, and matrix force maps. Using this data, we pretrain the TaF-Adapter to align tactile observations with ground-truth force signals in a shared latent space.
     (b) We fuse the TaF-Adapter into a VLA backbone and fine-tune the policy on real-world demonstrations enriched with force-aware language instructions (\ie, force-aware manipulation dataset).
     (c) This explicit tactile-force alignment empowers \modelname to master complex force-aware manipulation tasks, such as tool-use and deformable object manipulation, where traditional vision-based baselines consistently fail.}
    \label{fig:teaser}
\end{figure*}

\begin{abstract}
Vision-Language-Action (VLA) models have recently emerged as powerful generalists for robotic manipulation. However, due to their predominant reliance on visual modalities, they fundamentally lack the physical intuition required for contact-rich tasks that require precise force regulation and physical reasoning. 
Existing attempts to incorporate vision-based tactile sensing into VLA models typically treat tactile inputs as auxiliary visual textures, thereby overlooking the underlying correlation between surface deformation and interaction dynamics.
To bridge this gap, we propose a paradigm shift from tactile-vision alignment to tactile-force alignment. 
Here, we introduce TaF-VLA, a framework that explicitly grounds high-dimensional tactile observations in physical interaction forces. 
To facilitate this, we develop an automated tactile-force data acquisition device and curate the TaF-Dataset, comprising over 10 million synchronized tactile observations, 6-axis force/torque, and matrix force map. To align sequential tactile observations with interaction forces, the central component of our approach is the Tactile-Force Adapter (TaF-Adapter), a tactile sensor encoder that extracts discretized latent information for encoding tactile observations. This mechanism ensures that the learned representations capture history-dependent, noise-insensitive physical dynamics rather than static visual textures. Finally, we integrate this force-aligned encoder into a VLA backbone. Extensive real-world experiments demonstrate that TaF-VLA policy significantly outperforms state-of-the-art tactile-vision-aligned and vision-only baselines on contact-rich tasks, verifying its ability to achieve robust, force-aware manipulation through cross-modal physical reasoning.
\end{abstract}

\begin{IEEEkeywords}
    Foundation model, tactile and force sensing, force-aware manipulation.
\end{IEEEkeywords}

\section{Introduction}\label{sec:introduction}
Recent advances in \ac{vla} models have driven significant progress in robotic manipulation by jointly modeling large-scale visual representations and natural-language instructions~\cite{jiang2022vima,brohan2022rt,black2024pi0,kim2024openvla}. While these models demonstrate strong generalization across open-ended tasks, reliance on vision alone remains fundamentally insufficient for force-sensitive scenarios, such as handling fragile objects~\cite{zhu2025deformable,rtac}, precision assembly~\cite{zhu2018robot,zhao2025transferable}, or interaction with thin, deformable objects~\cite{lin2025pp}. Visual observations are susceptible to hand-object occlusion and inherently lack direct perception of interaction forces, which are critical for safe and robust execution~\cite{wen2020robust,hao2025tla,bi2025vla,yu2025demonstrating}. This sensory gap restricts current \ac{vla} approaches from attaining the precision and stability required for contact-rich manipulation, where feeling the interaction is as critical as seeing it~\cite{chen2022visuo,yu2025forcevla,DexMove_ICLR2026}.

To bridge the gap left by vision-only policies, prior research has largely relied on global force feedback, typically acquired via wrist-mounted Force/Torque (F/T) sensors~\cite{yu2025forcevla} or joint torque sensors~\cite{liu2025factr,zhang2025ta}. While these signals provide a useful measurement of total interaction load, they suffer from two critical shortcomings that limit their utility. First, from a hardware perspective, high-precision F/T sensors are expensive, fragile, and difficult to deploy at the scale required for great data-collection efforts, and are not broadly compatible with typical robot hardware~\cite{cao2021six}. Second, and more fundamentally, the data they provide is low-dimensional and spatially aggregated~\cite{lepora2021soft}. A single resultant force vector compresses complex contact interactions into a few scalar values, discarding the spatially distributed pressure patterns and local contact geometry that are essential for understanding how an object is being grasped or manipulated~\cite{calandra2017feeling}.

In contrast, \acp{vbts} have emerged as a low-cost, powerful multi-modal sensing element for capturing these missing local dynamics~\cite{luo2025tactile,calandra2018more}. Unlike resultant force vectors, these sensors can detect surface deformation~\cite{yuan2017GelSight,ma2019dense}, offering rich, high-resolution feedback on geometry, texture, local pressure distribution, and slippage. This visual nature makes them uniquely compatible with modern vision-based learning methods~\cite{lambeta2020digit}. By treating touch as an image, these sensors provide the dense, semantically rich information necessary to distinguish between stable grasps, slips, and intricate contact states, which are capabilities that global force sensors cannot achieve~\cite{yuan2015measurement,yuan2017GelSight}.

Recent efforts to incorporate tactile feedback into \ac{vla} architectures typically implant tactile streams as raw visual tokens, effectively aligning them with language and scene images~\cite{yu2024octopi,wu2025freetacman}. However, this strategy overlooks the unique modality of tactile sensing: unlike scene vision, which captures remote photometric properties, tactile sensing captures local mechanical interactions. Treating tactile data solely as ``more vision'' fails to extract the contact force information critical for force-aware manipulation tasks. \emph{We argue that to be effective, tactile representations must be grounded in the physical quantities they represent.} This motivates us to propose a paradigm shift from tactile-vision alignment to tactile-force alignment. By explicitly grounding tactile representations in physical force measurements, we equip the policy with a true understanding of contact dynamics.

To realize this paradigm shift, we propose \modelname, a \ac{vla} model architecture that utilizes high-dimensional visuotactile observations aligned with physical interaction forces. Developing such a system, however, presents three fundamental challenges:
\begin{enumerate}[label=(Q\arabic*), leftmargin=*]
\item \textbf{Scarcity of Tactile Data for Alignment:} \ac{vla} models require massive datasets to generalize. Existing datasets lack synchronized pairs of \textit{visuotactile and ground-truth force} data. How can we acquire such synchronized tactile-force alignment data at scale?
\item \textbf{Representation Disparity:} 
Visuotactile inputs are high-dimensional and geometric, whereas force signals are low-dimensional and dynamic. This modality gap raises a critical design choice: Should we rely on explicit force regression, which risks poor cross-sensor generalization? Or should we pursue implicit latent alignment? If the latter, the challenge lies in constructing a representation that captures history-dependent dynamics while maintaining robustness against sensor noise and hardware variance.
\item \textbf{Policy Integration:} From an architectural perspective, how do we fuse this tactile-force-aligned representation into large-scale \ac{vla} backbones, and without causing modality collapse or diluting the physical signal?
\end{enumerate}

To address these challenges, the development of \modelname comprises three key steps (see \cref{fig:teaser}).
\textit{First}, we design an automated data acquisition device that addresses the scarcity of paired tactile data and force data required for alignment. Each collected sample includes time-synchronized visuotactile observations, 6-axis force/torque measurements, and matrix force maps. The device adopts a parallel actuation structure that applies forces of the same magnitude and direction simultaneously to both the tactile sensor and the force sensors. By adjusting the stiffness and geometry of the replaceable indenters, we further enrich the diversity of contact interactions represented in the dataset. Thanks to its modular design, we can easily swap in different \acp{vbts}, and we collect data on six distinct sensors. The system operates efficiently, achieving a throughput of 100,000 synchronized tactile-force frames per hour. Using this setup, we construct the TaF-Dataset, a large-scale real-world tactile dataset comprising over 10 million data frames.

\textit{Second}, to bridge the critical representation gap between visuotactile data and force, we propose the Tactile-Force Adapter (TaF-Adapter). 
The TaF-Adapter employs contrastive learning to construct a shared latent space where tactile embeddings are aligned with paired force profiles. Compared to explicit and direct force calibration methods~\cite{helmut2024learning}, aligning only in the latent space without explicit reconstruction performs better when applied to new sensors (refer to experiments), which is crucial for practical deployment.
Moreover, we found that the mapping relationship from tactile images to contact forces is history-dependent, meaning that a single static frame cannot accurately represent dynamic force interactions. To capture such dynamic forces, the TaF-Adapter aggregates historical observations to improve expressibility. Moreover, because forces during manipulation are often noisy, the latent space is  quantized as a discrete codebook. These designs ensure that TaF-Adapter can capture physical dynamics robustly.

\textit{Finally}, we integrate this tactile-force-aligned module into a pretrained \ac{vla} backbone, leading to the design of \modelname model. By interleaving these tactile-force-aligned tokens into the language-action stream, the model learns to condition its end-effector actions on both language and tactile feedback cues. Benchmarking across 7 force-critical daily manipulation tasks demonstrates that \modelname achieves an average 22\% improvement in success rate over the state-of-the-art vision-tactile-aligned baseline, especially in contact-rich scenarios requiring precise force perception and regulation.

In summary, the contributions of this work are threefold:
\begin{itemize}[leftmargin=*]
\item \textbf{Tactile-Force Data Acquisition Device (address Q1):} We develop a low-cost, automated device and pipeline for collecting aligned tactile-force data at scale (\cref{sec:dataset}).
\item \textbf{Tactile-Force Alignment (address Q2):} We introduce the TaF-Adapter, a module that maps sequential tactile observations into a force-aligned latent space using a contrastive learning framework. By constructing a vector-quantized shared latent space that aligns temporal visuotactile data with 6-axis force/torque signals and matrix pressure maps, our approach learns representations that are robust to force noise and cross-sensor variation, while capturing rich, history-dependent contact dynamics.
(\cref{sec:align:encoder}).
\item \textbf{\modelname Policy (addressing Q3):} We developed the \modelname model, a VLA framework capable of incorporating tactile information. Experiments show that explicit force alignment enables \ac{vla} policies to successfully perform force-sensitive manipulation tasks that are otherwise intractable for vision-only or naive tactile-vision aligned baselines (\cref{sec:tafvla,sec:experiment}).
\end{itemize}

\section{Related Work}\label{sec:review}

\subsection{Vision-Language-Action Models and the Physical Gap}\label{sec:review:vla}

The paradigm of \ac{vla} learning has fundamentally transformed robotic manipulation by unifying large-scale visual representation learning with natural language reasoning~\cite{driess2023palm,gao2023k,luo2024multistage,Lin_2025,li2025controlvla,kawaharazuka2025vision}. Influential frameworks, ranging from transformer-based action predictors~\cite{brohan2022rt,zitkovich2023rt,team2024octo} to open-weight foundation models~\cite{kim2024openvla,black2024pi0,intelligence2025pi05}, leverage massive datasets of paired visual observations, language instructions, and action demonstrations~\cite{gu2023maniskill2,dasari2020robonet,walke2023bridgedata,liu2023libero}. Driven by the scaling laws of sequence modeling architectures, these systems have demonstrated remarkable generalization in complex contact-rich tasks such as grasping, manipulating articulated objects, cloth folding, and etc.~\cite{vaswani2017attention,ho2020denoising}.

However, a critical gap exists between \textit{semantic understanding} and \textit{physical interaction}. While current \acp{vla} excel at ``what to do'', they struggle with ``how to interact'' in contact-rich scenarios where visual feedback is insufficient~\cite{bi2025vla,zhang2025vtla}. This limitation is twofold. First, RGB cameras suffer from occlusion and lack the direct perception of contact forces, friction, and compliance required for tasks like precise insertion~\cite{zhang2025ta,hao2025tla} or in-hand manipulation~\cite{yousef2011tactile,andrychowicz2020learning,liuvtdexmanip,lidexdeform}. Second, standard robot datasets~\cite{o2024open,khazatsky2024droid} are dominated by high-level semantic labels (\eg, ``pick up the cup'') while omitting the fine-grained force directives (\eg, ``make a gentle press'', ``grasp tightly'', ``keep a compliant touch'') necessary to specify delicate interactions. Consequently, standard \ac{vla} policies are effectively ``force-blind'', relying on visual priors that often fail during complex physical interaction.

\subsection{Force Perception via Vision-based Tactile Sensing}\label{sec:review:vbts}
To retrieve these missing physical cues, \acfp{vbts} have emerged as a powerful solution, converting contact deformation into high-resolution images~\cite{luo2025tactile}. Accurately estimating contact forces from these images is a prerequisite for reliable control, and methodologies have evolved significantly in recent years~\cite{shahidzadeh2025feelanyforce,dong2017improved,yuan2017GelSight,lambeta2020digit}.
Early approaches relied on physics-driven calibration, inferring forces by tracking markers~\cite{mirzaee2024multiphysics} or solving inverse finite element methods (FEM) based on deformation fields~\cite{ma2019dense,li2020f,li20233,helmut2024learning}. While offering strong physical interpretability, these methods require precise knowledge of sensor geometry and are highly sensitive to material aging and manufacturing variability~\cite{dong2017improved,zhao2023situ}.
Consequently, recent works have pivoted toward data-driven approaches, employing neural networks to regress forces from labeled data collected via external force/torque sensors~\cite{lu20243d,chen2024generalizable}. More importantly, recognizing that contact is inherently dynamic, researchers have adopted temporal modeling (\eg, RNNs, Transformers) to capture history-dependent phenomena such as viscoelastic hysteresis, incipient slip, and stick-slip transitions~\cite{lin20239dtact,mandil2022action,fang2025force}.
These advances in temporal force estimation highlight a crucial insight: a tactile observation is not merely characterized by a static texture, but a snapshot of an evolving dynamical process. However, effectively transferring this insight into generalist \ac{vla} policies remains an open challenge.

\subsection{Tactile Integration in Generalist Policies}\label{sec:review:integration}
Recent efforts to integrate tactile sensing into \ac{vla} frameworks have met with mixed success. Some approaches augment policies with global force feedback via wrist-mounted F/T sensors~\cite{yu2025forcevla} or joint-torque estimates~\cite{liu2025factr,zhang2025ta}. While useful for collision detection, these signals aggregate interaction forces into a single vector, losing the spatial richness required to model local contact geometry.
Conversely, visuotactile-based \ac{vla} policies attempt to leverage the high resolution of vision-based tactile sensors~\cite{wu2025freetacman}. However, the predominant strategy is to treat tactile frames as auxiliary visual textures, aligning tactile embeddings with scene images using standard Vision Transformers (ViT). We argue that this represents yet another limitation in semantic representation: it aligns tactile data with \textit{visual appearance} rather than \textit{physical dynamics}, which are characterized by a temporal process. By neglecting the temporal dependencies identified in the force estimation literature (\cref{sec:review:vbts}), the current tactile-vision alignment methods fail to distinguish between static surface features and dynamic force events. Furthermore, without force signals as anchors, contrastive representation learning cannot theoretically guarantee that the learned embedding space is organized in a force-aware or force-consistent manner~\cite{ge2021robust,dave2024multimodal}.

\begin{figure*}[t!]
    \centering
    \includegraphics[width=\linewidth]{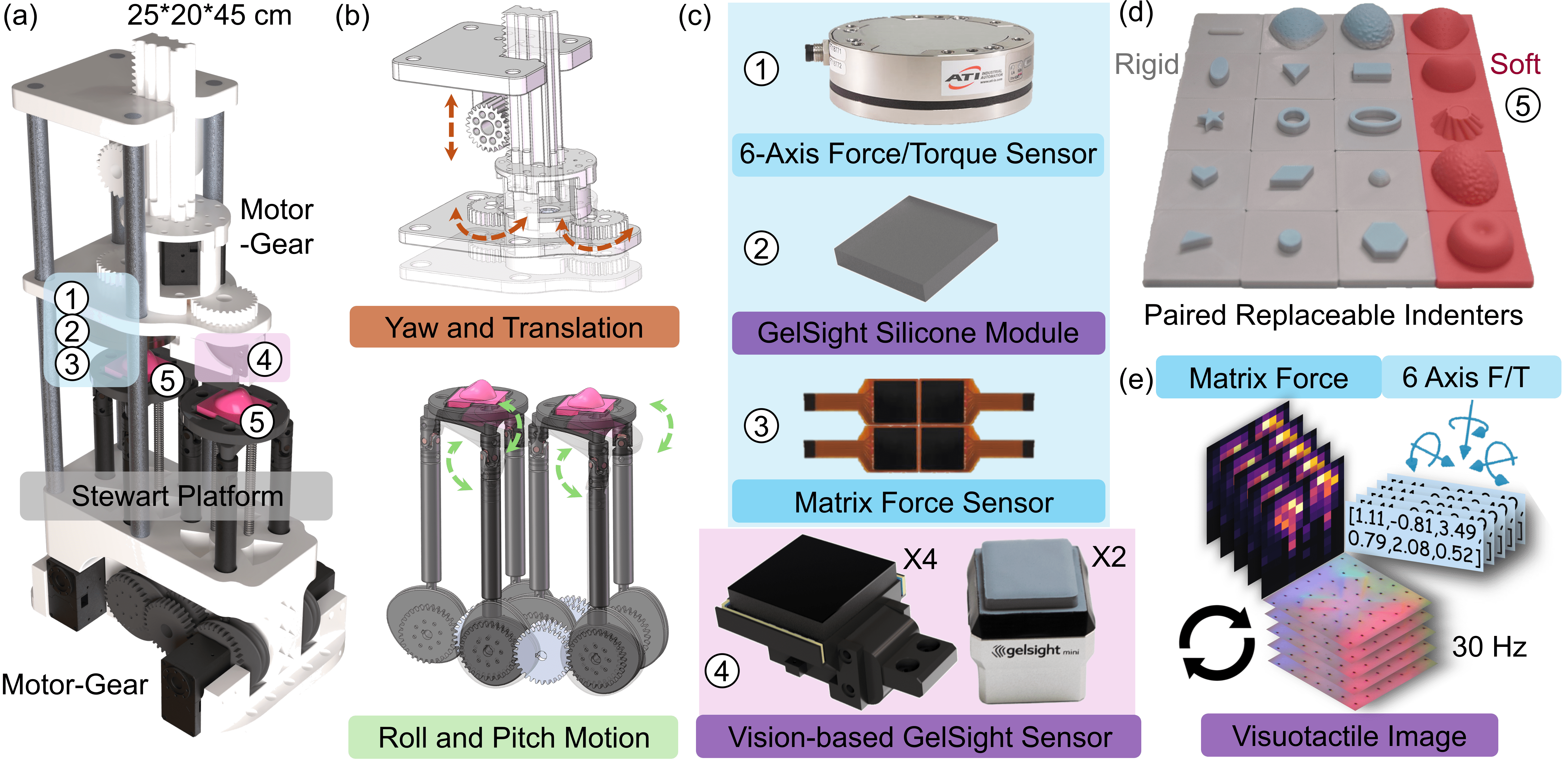}
    \caption{\textbf{The hardware design of the tactile-force data acquisition device that autonomously captures synchronized paired frames.} (a) Overall view of the device. (b) Kinematics of the platforms. (c) Force and tactile sensors that mounted on the system. In particular, the vision-based tactile sensor includes four self-made sensors (each silicone surface carrying a distinct marker pattern) and two GelSight Mini supplied in both marker-printed and marker-free versions. (d) Example contact indenter samples with varying patterns, curvatures, and hardness. (e) Synchronized tactile-force frames. }
    \label{fig:device}
\end{figure*}

\subsection{Distinction and Innovations of our \modelname}\label{sec:review:tafvla}
The advancement of \modelname lies in synthesizing these insights into a unified framework for generalist policy learning. Compared to the conventional ``tactile-vision alignment'' paradigm, our approach introduces three key innovations.
First, building on the dataset collected via our scalable acquisition pipeline, we propose a temporal adapter (TaF-Adapter) that fuses tactile observations across a receding horizon, enabling the policy to capture history-dependent dynamics such as incipient slip.
Second, instead of explicitly regressing forces (an approach that introduces additional force encoding steps via networks), we employ contrastive learning to align the tactile latent space with ground-truth force trajectories during pre-training. This yields a ``tactile-force'' aligned representation that encodes richer contact dynamics while eliminating the need for a costly F/T sensor at deployment.
Finally, we bridge the semantic gap by conditioning the policy on force-aware language instructions, allowing the model to perform precise, language-guided force modulation (a capability not commonly supported in current \ac{vla} models).

\section{TaF-Dataset}\label{sec:dataset}

The proposed tactile-force alignment paradigm hinges on the availability of high-fidelity, synchronized multi-modal data, specifically the visuotactile observations and ground-truth force/torque measurements. However, existing robotic datasets typically lack this precise alignment, often omitting dense force labels. To bridge this data gap, we present the TaF-Dataset, a large-scale repository designed to enable physically grounded representation learning. In this section, we first detail the automated hardware platform developed to acquire this data at scale (\cref{sec:dataset:device}). Subsequently, we describe the dataset construction process, including the automated acquisition protocol and the diversification strategies used to cover the full spectrum of contact dynamics (\cref{sec:dataset:dataset}).

\subsection{Tactile-Force Data Acquisition Device}\label{sec:dataset:device}

Our goal is the concurrent collection of 6-axis force/torque, matrix force map, and VBTS tactile data for alignment. Nevertheless, directly placing VBTS on top of a pressure sensor or an F/T sensor can alter the force transmission path and introduce significant distortion to the sensor readings. Additionally, scalable dataset construction demands highly efficient data acquisition. To overcome these limitations, we developed an autonomous data-collection platform that records synchronized high-resolution tactile images together with multi-dimensional force measurements, all with minimal human intervention. The design and working principle of the proposed mechanism are detailed below.

\subsubsection{Design Principle}
To collect multiple sensor responses concurrently under identical indentation conditions, we designed a machine with two key characteristics. The core principle is to generate highly synchronized indentation motions onto two twin platforms. Specifically:

\setstretch{1.03}

\textbf{Highly Accurate Synchronized Indentation:} To ensure that the tactile and force signals are precisely aligned in both time and space, we enforce strict kinematic coherence between the two platforms. Each corresponding \ac{dof} is driven by an identical motor-gear pair, mechanically linking the platforms and eliminating relative phase errors.

\textbf{Ease of Fabrication:} A second design goal is to enable broader replication by the research community. To this end, we adopted a modular architecture and prioritized accessibility. Aside from using standard off-the-shelf components (servomotors, springs, universal joints), the entire chassis is 3D-printed. We integrated self-aligning features to simplify assembly and used an open-frame lattice structure to preserve visibility during operation and installation. In addition, the modular, ``plug-and-play'' design supports flexible reconfiguration and customization for different contact scenarios.

\subsubsection{System Architecture}
As illustrated in \cref{fig:device}, the device comprises five integrated subsystems:

\textbf{Structural Chassis:}
Overall, the device features a compact footprint ($25\times 20\times 45$ cm) compatible with standard optical tables ($25 \times 25$ cm grid). The 3D-printed chassis is optimized for rapid assembly, allowing the full unit to be constructed in under four hours.

\textbf{Actuation Module:}
Dynamic motion is driven by five Dynamixel XM540 servomotors, peripherally mounted via quick-release couplings. The kinematic chain is split into two stages: two upper motors actuate the Z-axis translation and Yaw rotation, while three lower motors drive a parallel mechanism to generate Roll and Pitch inclinations. High-speed communication and control loops are managed via a U2D2 interface, ensuring precise trajectory tracking.

\setstretch{1.}

\textbf{Transmission Mechanism:}
To achieve synchronized motion of the two acquisition platforms, we employ a custom gear-driven transmission system. The lower stage utilizes a simplified Stewart platform configuration~\cite{dasgupta2000stewart} driven by cam-follower mechanisms. Specifically, three vertical columns are arranged radially at $120^{\circ}$ intervals. Vertical displacement of each column is driven by a rotating cam and constrained by a return spring; simultaneous differential actuation of these columns induces Roll (${\pm}15^{\circ}$) and Pitch (${\pm}15^{\circ}$) variations in the pressing plane.
The upper stage controls Yaw ($360^{\circ}$ continuous) and Z-axis translation via a cam-gear assembly. Crucially, the Z-axis mechanism includes a compliant element to maintain constant contact pressure between the indenter and the sensor, preventing overload during high-speed operation. 

\textbf{Contact Indenters:}
Both the upper and lower platforms feature quick-release grooves to facilitate the rapid exchange of contact indenters. We fabricated a library of 3D-printed samples featuring diverse surface patterns, curvatures, and stiffness levels (see \cref{fig:device}(d)). These samples are mounted in opposing pairs to simulate varied interaction scenarios, ranging from point contacts to conformal surface patches.

\textbf{Force and Tactile Sensors:}
The system integrates three distinct sensing modalities:
\begin{itemize}[leftmargin=*]
    \item
    \label{vbts_detail}
    \textbf{\ac{vbts}:} The Device under test. This tactile sensor captures high-resolution surface deformation at 30 Hz. In this work, we employed four custom-built sensors (marker-free and featuring 4×4, 7×7, and 9×9 densities) alongside two GelSight Mini sensors, one marker-free and the other with a 7×9 marker pattern. The silicone pads of all sensors share identical hardness.
    \item \textbf{6-Axis Force/Torque Sensing:} An ATI Axia80 sensor measures the 6D wrench at the contact interface.
    \item \textbf{Matrix Force Sensor:} A calibrated piezoresistive array ($12\times12$ taxels) provides spatial force distribution.
\end{itemize}

\begin{figure*}[t!]
    \centering
    \includegraphics[width=\linewidth]{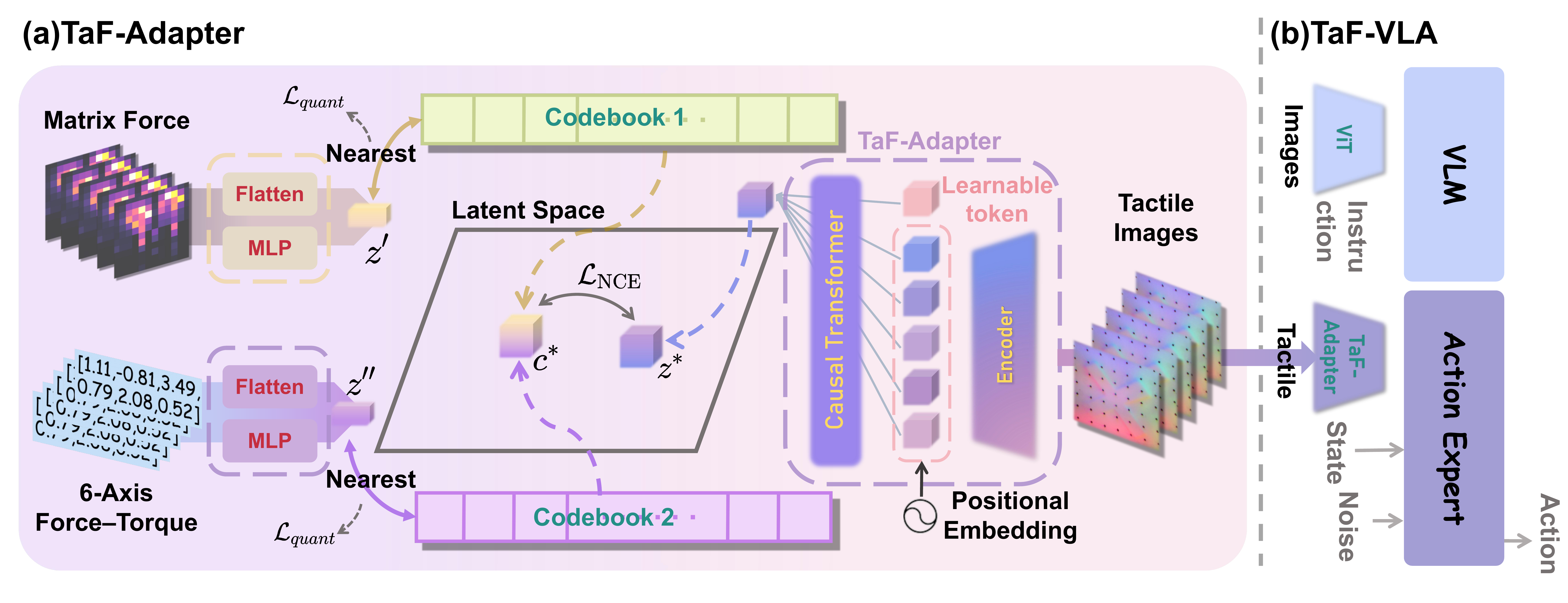}
    \caption{\textbf{Overview of the \modelname learning framework.} The architecture consists of two distinct stages: (a) Tactile-Force Alignment (TaF-Adapter): We propose a self-supervised learning paradigm. The Force Encoder ($f_{\phi'}, f_{\phi''}$) quantizes heterogeneous force signals, distributed pressure maps, and 6-axis F/T vectors, into two discrete codebooks via \ac{vqvae}, creating stable physical anchors. The Tactile Encoder ($f_\textrm{causal-TF}$) employs a Causal Transformer to process sequential tactile images. These branches are synchronized in a shared latent space via a contrastive InfoNCE objective ($\mathcal{L}_\text{NCE}$), effectively teaching the tactile encoder to infer force dynamics from visual deformation. (b) Policy Integration (\modelname): The pre-trained, frozen TaF-Adapter is integrated into a \ac{vla} backbone. It injects force-aligned tactile tokens alongside visual and language embeddings, allowing the Action Expert to generate force-aware manipulation trajectories.}
    \label{fig:pipeline}
\end{figure*}

All sensors are connected to the PC via either USB or Ethernet cables. Given that the force sensors operate at frequencies much faster than \ac{vbts}, data collection was synchronized to the 30 Hz of the visuotactile sensor via resampling to enable paired data acquisition.

\subsection{TaF-Dataset Construction}\label{sec:dataset:dataset}
We leverage the hardware platform to construct athe TaF-Dataset, which is designed to establish a mapping between \ac{vbts}'s visual output and corresponding physical force measurements. To achieve this, one \ac{vbts} sensor is suspended above the right platform, while a 6-axis force/torque sensor, a silicone piece, and a matrix force sensor are mounted in sequence above the left platform. The two elastomer surfaces are manually aligned in height before data collection. For matrix force regions that fall outside the GelSight Mini, we pad them with zeros. Then, two identical indenters are placed on both platforms. This configuration ensures that the tactile and force sensors are subjected to forces of identical magnitude and direction, thereby yielding highly aligned multimodal data.
Moreover, the system can record data while applying forces of varying magnitudes and directions to the sensors through indenters, enabling the collection of over 100K synchronized frames per hour and effectively overcoming the scalability limitations of manual data acquisition.

\subsubsection{Automated Generation Protocol}
The collection process requires human intervention only for swapping indenter sets and different visuotactile sensors. As illustrated in \cref{fig:device}(b), the cycle begins by driving the Z-axis to establish contact between the sensor and the indenter. Once contact is made, the system executes a ``press-and-perturb'' strategy: it dynamically varies the Z-axis pressure while simultaneously driving the Pitch, Roll, and Yaw motors. This induces continuous and diverse trajectories that explore the full state space of the sensor, ranging from static indentation to incipient slip, and far more efficiently than discrete probing methods.

\subsubsection{Contact Diversity}
To ensure the learned representations are robust to the variability of the real world, we explicitly maximize diversity following three strategies: 1) Rather than using fixed position control, we modulate the torque of the Z-axis actuator. This generates temporally continuous contact pressures, allowing the acquisition of sensor response from the lightest touch to deep saturation. 2) By randomizing the rotational velocities of the parallel platform, the system introduces intricate shear forces and torsional moments. This ensures the dataset contains not just normal force data, but also the friction-dominated signals critical for manipulation stability. 3) We employ a diverse library of over 60 distinct indenters, comprising 40 rigid and 20 deformable samples. These range from standard geometric primitives to complex everyday objects with varying curvatures and hardness. This physical variety forces the model to learn generalized deformation features rather than overfitting to specific object shapes.

\subsubsection{Dataset Scale and Statistics}
The resulting TaF-Dataset is among the largest multi-modal tactile repositories available. To support the cross-sensor generalization claims in this work, we aggregated data across 6 distinct \ac{vbts} units, including 4 custom-built variants and 2 commercial GelSight sensors (see \cref{vbts_detail}). In total, the dataset comprises over 10 million synchronized tactile-force pairs, providing the high-fidelity supervision necessary to align the visual and force domains.

\section{Tactile-Force Alignment and VLA Integration}\label{sec:align}
To leverage the collected dataset and integrate force perception into VLA models, we propose the TaF-Adapter. We depart from prior approaches such as direct force regression~\cite{helmut2024learning} which fall short in cross-sensor generalization, and tactile-vision alignment~\cite{yu2024octopi,feng2025anytouch} which captures surface textures but neglects physical interaction forces. Instead, the TaF-Adapter employs an implicit alignment strategy, mapping tactile sequences and force signals into a shared latent space.

To capture the dynamic nature of force, the TaF-Adapter employs historical observations, aligning tactile and force sequences via contrastive learning. Furthermore, to mitigate high-frequency sensor noise, we apply vector quantization, which discretizes the latent space into separate codebooks for 6-axis force/torque measurements and matrix-structured force maps. This design ensures the learning of stable, cross-sensor representations explicitly grounded in physical dynamics.

We instantiate the \modelname policy by integrating the TaF-Adapter into a pretrained \ac{vla} backbone \cite{intelligence2025pi05}. By interleaving force-aligned tactile tokens into the language-action stream, the policy conditions its execution on both semantic instructions and tactile feedback, enabling robust manipulation in contact-rich scenarios.

\subsection{TaF-Adapter: Tactile-Force Alignment }\label{sec:align:encoder}
The training architecture for TaF-Adapter (illustrated in \cref{fig:pipeline}(a)) consists of two encoding procedures. First, the force encoder (left) quantizes sequential force signals into a discrete codebook, capturing dynamic force primitives. Second, the tactile encoder (right) encodes temporal sequences of tactile images and is trained to map these visual inputs to their corresponding force tokens via contrastive learning. This design effectively ``teaches'' the model to perceive the interaction force from \ac{vbts}.

\paragraph{Force Quantization}\label{sec:align:force_encoder}
Force and pressure signals are typically high-bandwidth and exhibit noise with modality-dependent characteristics. Similar challenges have been widely studied in the speech recognition and action-generation communities, where discretized latent representations from \ac{vqvae}~\cite{VQVAEvan2017neural} are used to mitigate noise and capture reusable dynamics \cite{oord2018neural,baevski2020vqwav,lee2023vector, sadok2023vector}. Following this idea, we encode force within a continuous time window using a learnable \ac{vqvae} codebook. The resulting discretized tokens represent essential patterns of force primitives while providing compactness and semantic interpretability.

Let the raw force measurement at time step $i$ be defined as a tuple $F_i = (F'_i, F''_i)$, where:
\begin{itemize}[leftmargin=*]
    \item $F'_i \in \mathbb{R}^{12 \times 12}$ represents the spatial pressure map (normal force distribution).
    \item $F''_i \in \mathbb{R}^{6}$ represents the 6-axis force/torque vector (global wrench).
\end{itemize}

To capture temporal information, we operate on a sliding window of length $N$\footnote{Empirically set to 5 in this paper.}. We employ two dedicated MLP encoders to project these sequences into latent space:
\begin{align}
    z' &= f_{\phi'}(\{F'_{i+j}\}_{j=0}^{N-1}), \\
    z'' &= f_{\phi''}(\{F''_{i+j}\}_{j=0}^{N-1}).
\end{align}

We define two learnable codebooks, $\mathcal{C}' = \{c'_k\}_{k=1}^{K'}$ and $\mathcal{C}'' = \{c''_m\}_{m=1}^{K''}$, corresponding to the pressure and wrench modalities, respectively. Each latent vector is quantized to its nearest neighbor in the codebook:
\begin{align}
    (c')^{*} &= \arg\min_{c \in \mathcal{C}'} \|z' - c\|_2, \\
    (c'')^{*} &= \arg\min_{c \in \mathcal{C}''} \|z'' - c\|_2.
\end{align}
The final force embedding $c^{*}$ is the concatenation of these quantized tokens:
\begin{equation}
    c^{*} = \text{concat}\big((c')^{*}, (c'')^{*}\big) \in \mathbb{R}^{d}.
\end{equation}

The network is trained using the standard \ac{vqvae}~\cite{VQVAEvan2017neural} objective, combining reconstruction loss with the quantization loss:
\begin{small}
\begin{align}
    \mathcal{L}_{\mathrm{quant}} = \sum_{x \in \{', ''\}} \Big( \|\text{sg}[z^x] - (c^x)^{*}\|_2^2 + \beta \|z^x - \text{sg}[(c^x)^{*}]\|_2^2 \Big),
\end{align}
\end{small}
where $\text{sg}[\cdot]$ denotes the stop-gradient operator and $\beta$ is the commitment weight.

Crucially, relying solely on $\mathcal{L}_{\mathrm{quant}}$ may lead to \textit{codebook collapse}~\cite{razavi2019generating}, a degenerate scenario where the encoder maps diverse inputs to a trivial subset of the codebook. To explicitly prevent this and ensure the discrete tokens preserve high-fidelity physical semantics, we introduce a reconstruction mechanism. We employ decoders $g_{\psi'}$ and $g_{\psi''}$ to recover the original signal sequences from the quantized embeddings. The total force learning objective combines the reconstruction loss with the quantization regularization:
\begin{equation}
    \mathcal{L}_{\mathrm{recon}} = \sum_{x \in \{', ''\}} \|\{F^x_{i+j}\} - g_{\psi^x}((c^x)^{*})\|_2^2.
\end{equation}
By enforcing precise signal reconstruction, the codebook is compelled to span the variance of the force dynamics, thereby serving as a robust anchor for the subsequent tactile alignment.

\begin{figure*}[t!]
    \centering
    \includegraphics[width=\linewidth]{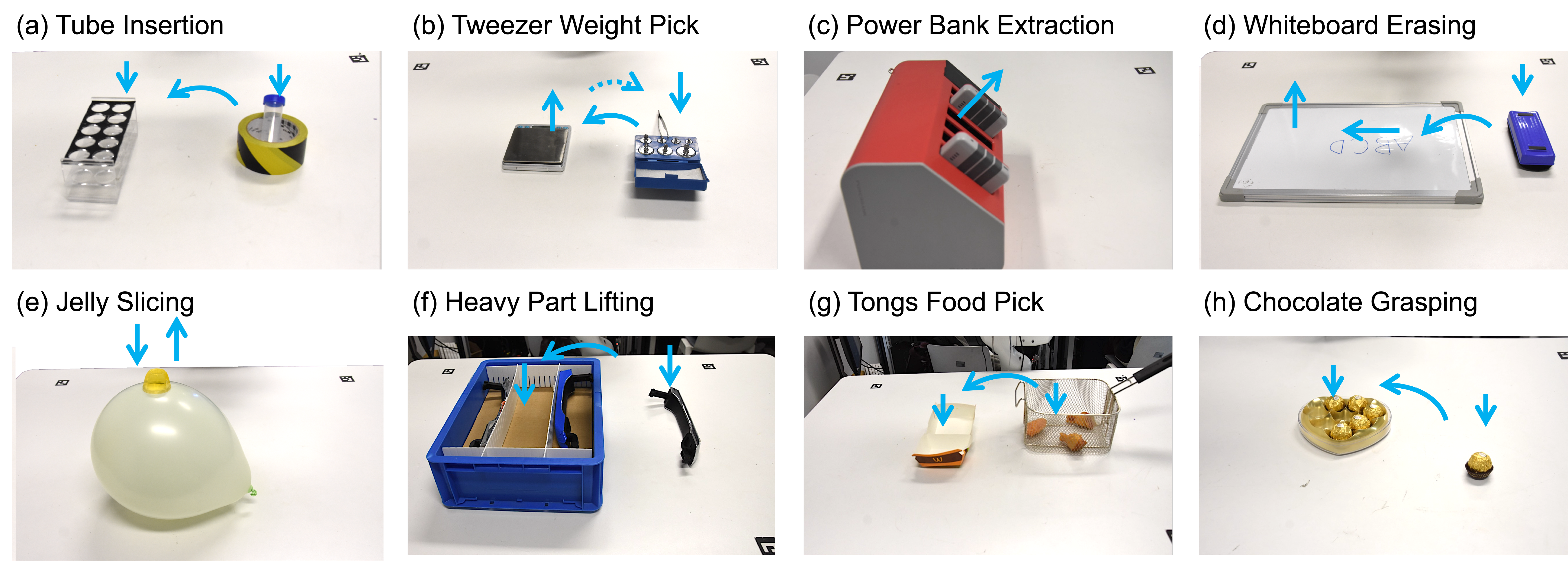}
    \caption{Overview of the eight force-aware manipulation tasks. Blue arrows indicate end-effector trajectories.} 
    \label{fig:setup}
\end{figure*}

\paragraph{Temporal Tactile Encoding}\label{sec:align:tactile_encoder}
The tactile encoder is responsible for inferring latent feature from sensor's visual output. Given a synchronized sequence of tactile images $\{ I^{\mathrm{tac}}_{i+j} \}_{j=0}^{N-1}$, we first extract per-frame features using a Vision Transformer (ViT) \cite{dosovitskiy2020image} backbone:
\begin{equation}
    \{ h^{\mathrm{tac}}_{i+j} \}_{j=0}^{N-1} = f_{\mathrm{ViT}}\big(\{ I^{\mathrm{tac}}_{i+j} \}_{j=0}^{N-1}\big).
\end{equation}
To aggregate temporal information, which is critical for distinguishing static deformation from slip or hysteresis, we append a learnable summary token $s$ and process the sequence with a causal Transformer encoder:
\begin{equation}
    z^{*} = f_{\mathrm{causal-TF}}\big(s, \{h^{\mathrm{tac}}_{i+j}\}_{j=0}^{N-1}\big).
    \label{eq:tactile_inference}
\end{equation}
The output embedding corresponding to $s$ serves as the holistic representation of the tactile event.

\paragraph{Cross-Modal Alignment}\label{sec:align:contrastive}
To align the tactile representation $z^{*}$ with the physical force code $c^{*}$, we employ a symmetric InfoNCE objective~\cite{oord2018representation}. This contrastive framework enforces that the tactile embedding of a specific interaction event is maximally similar to its ground-truth force code, while being dissimilar to force codes from other time steps.

For a batch of $B$ synchronized pairs $\{(z^{*}_b, c^{*}_b)\}_{b=1}^B$, the alignment loss is defined as:
\begin{equation}
    \mathcal{L}_{\mathrm{NCE}} = - \frac{1}{B} \sum_{b=1}^{B} \log \frac{\exp(\text{sim}(z^{*}_b, c^{*}_b)/\tau)}{\sum_{k=1}^{B} \exp(\text{sim}(z^{*}_b, c^{*}_k)/\tau)},
\end{equation}
where $\text{sim}(u, v) = u^\top v / (\|u\| \|v\|)$ is the cosine similarity and $\tau$ is a temperature hyperparameter. The total training objective for the TaF-Adapter is a weighted sum of the alignment and quantization losses:
\begin{equation}
    \mathcal{L}_{\mathrm{total}} = \mathcal{L}_{\mathrm{NCE}} + \lambda_1 \mathcal{L}_{\mathrm{quant}} + \lambda_2\mathcal{L}_{\mathrm{recon}}.
\end{equation}

By adopting a sliding-window temporal sampling strategy, the model learns the temporal evolution of contact forces directly from sequential observations. As a result, temporally adjacent frames exhibit naturally emerging similarity in the latent space, eliminating the need for manually designed distance metrics or explicit adjacency constraints.

\subsection{Force-aware VLA Policy Integration}\label{sec:tafvla}
We integrate the pre-trained TaF-Adapter into a state-of-the-art VLA backbone to instantiate the \modelname policy. We build upon the $\pi_{0.5}$ architecture~\cite{intelligence2025pi05}, a flow-matching-based generalist policy.

Given a third-person image $I_{\mathrm{global}}$, a wrist-mounted image $I_{\mathrm{wrist}}$, and a language instruction $l$, the encoder outputs a unified conditioning vector:
\begin{align}
\varphi = \pi_{\mathrm{VLM}}\big(I_{\mathrm{global}}, I_{\mathrm{wrist}}, l\big),
\end{align}
which provides high-level task semantics and global scene context for action generation.

To integrate tactile feedback, we fuse TaF-Adapter features with $\varphi$. The adapter extracts local contact cues from a sliding window of VBTS tactile signals (see \cref{eq:tactile_inference}). The tactile-conditioned flow-matching objective is:

\begin{small}
\begin{align}
\mathcal{L}_{\mathrm{FM}}
=
\mathbb{E}_{t,\,\tau,\,u}
\left[
\left\|
\pi_{\theta}\Big(
a^\tau_t, \tau \,\Big|\, \varphi_t, z^{\mathrm{tac}}_t, q_t
\Big)
-
\big(
a_t - u
\big)
\right\|^2_2
\right],
\label{eq:flowmatch}
\end{align}
\end{small}
where $a_t$ is the ground-truth action, $u\!\sim\!\mathcal{N}(0,I)$ is noise, and $a^\tau_t = \tau a_t + (1-\tau)u$ follows standard conditional flow-matching. Here, $q_t$ denotes robot proprioception. The policy $\pi_\theta$ learns a velocity field that drives $a^\tau_t$ toward $a_t$ under conditioning from $\varphi_t, z^{\mathrm{tac}}_t,$ and $q_t$.

During inference, \modelname generates an action by integrating the learned velocity field from noise to a clean action. Starting from $A_t^{0}\!\sim\!\mathcal{N}(0,I)$, the action estimate is updated as:
\begin{align}
A_t^{\tau+\delta}
=
A_t^{\tau}
+
\delta \,
v_{\theta}\!\left(
A_t^{\tau},
\tau \,\big|\, 
\varphi_t, z^{\mathrm{tac}}_t, q_t
\right),
\label{eq:inference_flow}
\end{align}
where $v_{\theta}$ is the learned velocity field and $\delta=0.1$ (following $\pi_{0.5}$). Integrating $\tau\!\in\![0,1]$ yields a noise-free action $A_t^{1}$ executed by the robot.

By conditioning on the global visual–language features $\varphi$ and the local tactile latent $z^{\mathrm{tac}}_t$, \modelname generates actions that remain both controllable and contact-aware. This design enables real-time adjustment of manipulation behaviors based on tactile feedback while consistently following high-level task directives.

\section{Experiments}\label{sec:experiment}
To evaluate the efficacy of the proposed \modelname framework and the underlying TaF-Adapter, we conduct a comprehensive suite of real-world, contact-rich manipulation experiments. Our analysis addresses four primary research questions:
\begin{enumerate}[label=({E\arabic*}),leftmargin=*]
\item \textbf{Performance Benchmarking:} Does the proposed explicit tactile-force alignment (TaF-Adapter) allow \modelname to outperform state-of-the-art vision-only and tactile-vision aligned baselines on diverse contact-rich tasks? (\cref{sec:exp:main_result})
\item \textbf{Generalization \& Robustness:} Can the TaF-Adapter generalize across different sensor hardware, and does its ``plug-and-play'' nature benefit other policy architectures? (\cref{sec:exp:cross_sensor})
\item \textbf{Training Efficiency:} Does integrating multi-modal feedback improve sample efficiency, allowing the model to outperform vision-only baselines with fewer demonstrations? (\cref{sec:exp:train_data})
\item \textbf{Ablation Studies:} What is the impact of specific design choices, such as using historical observations and discrete vector quantization, on system performance? (\cref{sec:exp:abalation})
\end{enumerate}

\begin{table*}[t]
\centering
\caption{Quantitative comparison of task success rates (\%) across seven manipulation tasks for all evaluated methods. Bold numbers denote the best-performing policy, while underlined numbers indicate the second-best-performing policy. }
\label{tab:exp_results}
\vspace{2mm} 

\resizebox{\textwidth}{!}{%
\begin{tabular}{lcccccccc}
\toprule
\multirow{2.5}{*}{\textbf{Task}} 
& \multicolumn{3}{c}{\cellcolor{actbg}\textbf{ACT Backbone}} 
& \multicolumn{2}{c}{\cellcolor{dpbg}\textbf{Diffusion Backbone}} 
& \multicolumn{2}{c}{\cellcolor{vlabg}\textbf{VLA($\boldsymbol{\pi}$) Backbone}} \\

\cmidrule(lr){2-4} \cmidrule(lr){5-6} \cmidrule(lr){7-8}

& \cellcolor{actbg}ACT 
& \cellcolor{actbg}FreeTacMan 
& \cellcolor{actbg}+TaF-Adapter 
& \cellcolor{dpbg}DP 
& \cellcolor{dpbg}+TaF-Adapter 
& \cellcolor{vlabg}$\pi_{0.5}$ 
& \cellcolor{vlabg}TaF-VLA (Ours) \\
\midrule

Tube Insertion   
& \cellcolor{actbg}26.7 & \cellcolor{actbg}\underline{53.3} & \cellcolor{actbg}46.7             
& \cellcolor{dpbg}33.3 & \cellcolor{dpbg}46.7             
& \cellcolor{vlabg}\underline{53.3} & \cellcolor{vlabg}\textbf{66.7} \\

Tweezer Weight Pick      
& \cellcolor{actbg}0.0  & \cellcolor{actbg}6.7  & \cellcolor{actbg}\underline{13.3} 
& \cellcolor{dpbg}0.0  & \cellcolor{dpbg}6.7              
& \cellcolor{vlabg}0.0              & \cellcolor{vlabg}\textbf{33.3} \\

Power Bank Extraction   
& \cellcolor{actbg}26.7 & \cellcolor{actbg}33.3 & \cellcolor{actbg}\underline{53.3} 
& \cellcolor{dpbg}20.0 & \cellcolor{dpbg}40.0             
& \cellcolor{vlabg}33.3             & \cellcolor{vlabg}\textbf{66.7} \\

Whiteboard Erasing 
& \cellcolor{actbg}13.3 & \cellcolor{actbg}26.7 & \cellcolor{actbg}\underline{33.3} 
& \cellcolor{dpbg}20.0 & \cellcolor{dpbg}\underline{33.3} 
& \cellcolor{vlabg}\underline{33.3} & \cellcolor{vlabg}\textbf{60.0} \\

Jelly Slicing        
& \cellcolor{actbg}13.3 & \cellcolor{actbg}26.7 & \cellcolor{actbg}26.7             
& \cellcolor{dpbg}13.3 & \cellcolor{dpbg}\underline{33.3} 
& \cellcolor{vlabg}13.3             & \cellcolor{vlabg}\textbf{53.3} \\

Heavy Part Lifting   
& \cellcolor{actbg}40.0 & \cellcolor{actbg}66.7 & \cellcolor{actbg}\underline{73.3} 
& \cellcolor{dpbg}46.7 & \cellcolor{dpbg}66.7             
& \cellcolor{vlabg}53.3             & \cellcolor{vlabg}\textbf{80.0} \\

Chocolate Grasping   
& \cellcolor{actbg}66.7 & \cellcolor{actbg}\underline{86.7} & \cellcolor{actbg}\underline{86.7} 
& \cellcolor{dpbg}73.3 & \cellcolor{dpbg}\underline{86.7} 
& \cellcolor{vlabg}73.3             & \cellcolor{vlabg}\textbf{93.3} \\

\midrule

\textbf{Average Success}  
& \cellcolor{actbg}26.7 & \cellcolor{actbg}42.8 & \cellcolor{actbg}\underline{47.6} 
& \cellcolor{dpbg}29.5 & \cellcolor{dpbg}44.7             
& \cellcolor{vlabg}37.1             & \cellcolor{vlabg}\textbf{64.8} \\

\bottomrule
\end{tabular}%
}
\end{table*}

\subsection{Experimental Setup}\label{sec:exp:setup}

\subsubsection{Training and Evaluation Settings}
For policy training, we curated a dataset comprising over 10,000 force-aware episodes across more than 20 diverse scenarios, ranging from fragile object handling to complex tool use (see \textcolor{red}{Appendix}~\ref{appendix:manipdataset}). For quantitative evaluation, we defined a targeted suite of 8 real-world, contact-rich tasks, as shown in \cref{fig:setup}. These tasks were specifically selected to span different types of physical interactions, aiming to comprehensively evaluate the model's force-reasoning capabilities. A detailed breakdown of the protocols and success criteria is provided in \textcolor{red}{Appendix}~\ref{appendix:task_analysis}.

\subsubsection{Baselines}
We compare \modelname against five \emph{open-source} baselines representing the current state-of-the-art:
\begin{itemize}[leftmargin=*]
    \item \textbf{\ac{act}}~\cite{zhao2023learning}: A standard Action Chunking with Transformers policy that relies solely on RGB images and proprioception.
    \item \textbf{FreeTacMan}~\cite{wu2025freetacman}: A representative \textit{vision–tactile alignment} baseline implemented on the \ac{act} architecture. It treats tactile feedback as auxiliary visual tokens, lacking explicit physical grounding.
    \item \textbf{\ac{act} + TaF-Adapter}: The standard \ac{act} policy augmented with our pre-trained TaF-Adapter, validating the benefit of tactile-force alignment over tactile-vision alignment (FreeTacMan) on the same backbone.
    \item \textbf{Diffusion Policy (DP)}~\cite{chi2025diffusion}: A generative behavior cloning policy based on denoising diffusion, trained using only RGB images and proprioception.
    \item \textbf{DP + TaF-Adapter}: The standard Diffusion Policy augmented with our TaF-Adapter, demonstrating the adapter's generalization to generative frameworks.
    \item $\boldsymbol{\pi_{0.5}}$~\cite{intelligence2025pi05}: A state-of-the-art \ac{vla} foundation model fine-tuned on our dataset using only visual observations, serving as the vision-only backbone for our proposed \modelname.
\end{itemize}

All models (including baselines and \modelname) are trained using the same dataset, and all real-world evaluations run with inference on an NVIDIA RTX 4090 workstation. Task performance is measured using \emph{success rate} (the criteria are described in \textcolor{red}{Appendix}~\ref{appendix:task_analysis}). Each model is given 15 rollout trials per task, and the success rate is computed as the proportion of successful attempts. Success criteria are task-dependent and represented as binary completion conditions (whether the task is successfully completed). 

\subsection{Comparison with Baselines}\label{sec:exp:main_result}
To highlight the advantages of our force-aware policy, we benchmarked it against several baselines on the aforementioned tasks. \cref{tab:exp_results} summarizes the quantitative results. Specifically, \modelname achieves the highest success rate across all scenarios. The performance advantage is most pronounced in force-critical tasks, where \modelname outperforms the strongest baseline by a substantial margin.

\textbf{Comparison to Vision-only Policies:}
Overall, performance advantages are evident for \modelname compared to vision-only policies, as shown in \cref{tab:exp_results}. Furthermore, we observed that task performance varies substantially depending on the task content. In tasks dominated by geometric constraints, such as \textit{Tube Insertion} where the initial alignment is paramount, vision-only baselines (\eg, $\pi_{0.5}$) already perform well ($53.3\%$), confirming that visual servoing is sufficient when spatial alignment is the primary objective. In these scenarios, the addition of tactile sensing yields only marginal gains.
However, the disparity becomes evident in tasks requiring dynamic force modulation. For heavier or friction-dependent objects, such as \textit{Tweezer Weight Pick} and \textit{Power Bank Extraction}, vision-only baselines frequently fail (\eg, DP, ACT, and $\pi_{0.5}$ all achieve $0\%$ on \textit{Tweezer Weight Pick}). We observed that beyond pre-contact errors, these policies fail to apply sufficient grasp force to prevent incipient slip. This occurs because micro-slip cues are visually imperceptible until the object has already been dropped.
Similarly, in \textit{Whiteboard Erasing} and \textit{Jelly Slicing}, vision-only policies struggle to regulate contact pressure, leading to either trajectory drift or substrate damage. \modelname excels in these regimes, achieving $60.0\%$ on \textit{Whiteboard Erasing} and $53.3\%$ on \textit{Jelly Slicing}. By grounding tactile feedback in physical force, the policy detects contact transitions and shear resistance successfully, enabling stable closed-loop control.

\begin{figure*}[t!]
    \centering
    \includegraphics[width=\linewidth]{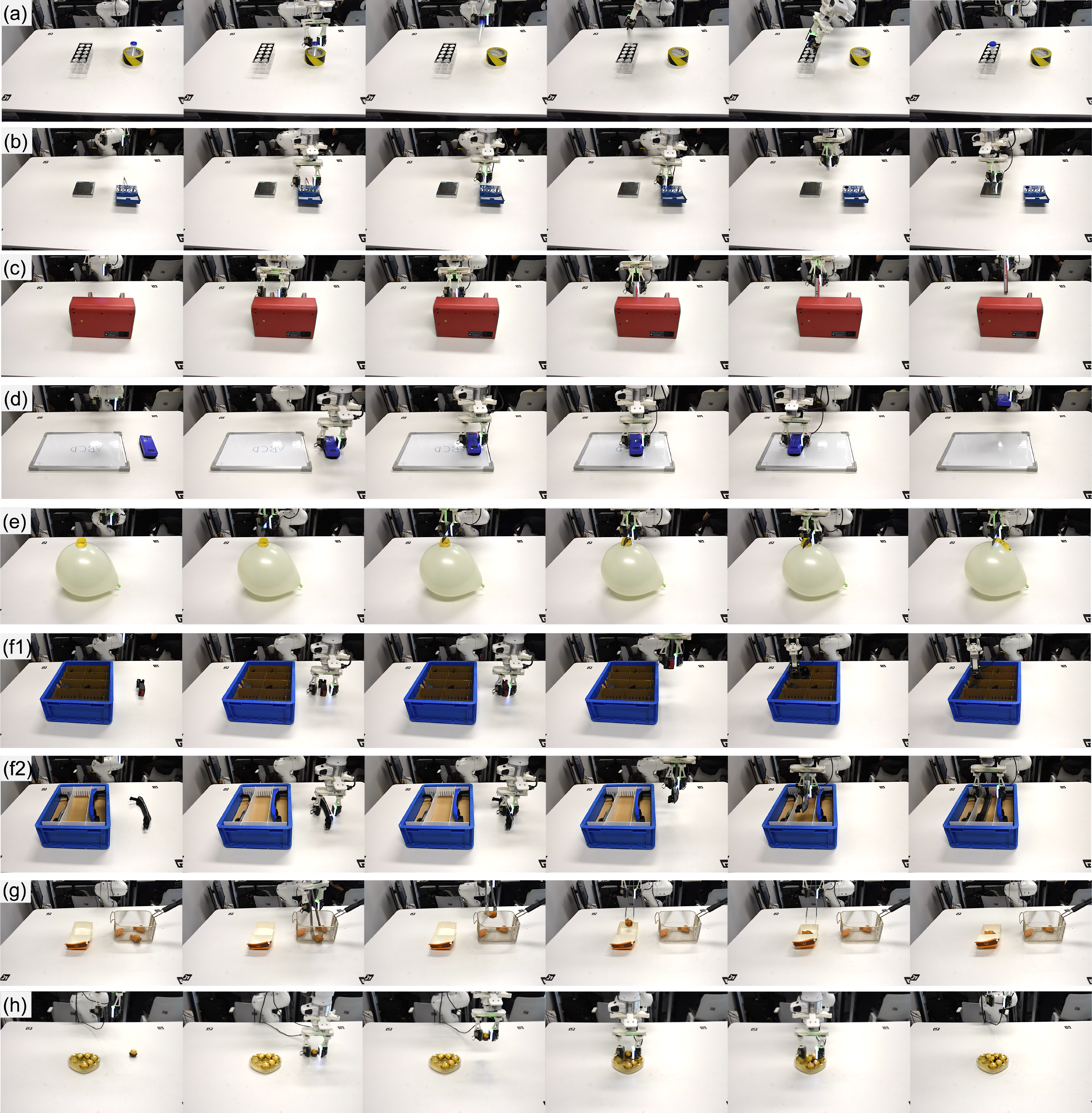}
    \caption{\textbf{Successful execution sequences across the evaluated task suite:} (a) Tube Insertion; (b) Tweezer Weight Pick; (c) Power Bank Extraction; (d) Whiteboard Erasing; (e) Jelly Slicing; (f1) Heavy Part Lifting (Buckle); (f2) Heavy Part Lifting (Door Handle); (h) Tongs Food Pick; (i) Chocolate Grasping.}
    \label{fig:gallery}
\end{figure*}

\textbf{Impact of Tactile-Force Alignment:} We compared \modelname with other tactile-based policies to evaluate the effect of our tactile-force alignment design. The results show that \modelname outperforms FreeTacMan by a substantial margin (\eg, +26.6\% on \textit{Tweezer Weight Pick} and +33.3\% on \textit{Whiteboard Erasing}). These findings collectively validate our core hypothesis: treating tactile data merely as ``auxiliary vision'' is insufficient for force-aware tasks. While FreeTacMan captures the \emph{appearance} of contact, it lacks the calibrated understanding of \emph{magnitude} necessary to regulate interactions. In contrast, by grounding tactile embeddings in physical force and pressure, \modelname successfully handles force-sensitive tasks, such as detecting stiffness boundaries in \textit{Jelly Slicing} and managing binding friction in \textit{Power Bank Extraction} with significantly higher success rates. Visualizations of successful rollouts are shown in \cref{fig:gallery}.

\textbf{Plug-and-Play Capability:}
A key finding is that simply attaching the TaF-Adapter to an existing policy yields immediate performance gains. We evaluated the \textit{ACT + TaF-Adapter} and \textit{DP + TaF-Adapter} variants, both of which consistently outperform their vanilla counterparts across all tasks, with absolute improvements ranging from 6.7\% to 33.3\%. For example, on \textit{Jelly Slicing}, the success rate of DP increases from 13.3\% to 33.3\% when augmented with the TaF-Adapter. Interestingly, when the standard \ac{act} baseline is equipped with our TaF-Adapter, it surpasses its previous tactile-enhanced variant FreeTacMan in force-critical scenarios (\eg, +20.0\% on \textit{Power Bank Extraction}). These results demonstrate that the TaF-Adapter functions as a general-purpose module capable of injecting physical tactile intuition into standard visuomotor policies.

\begin{table*}[t]
    \centering
    \caption{Performance (success rates, \%) comparison of different tactile encoding methods across sensors. 
    ``Seen'' denotes sensors used during training; ``Unseen'' denotes out-of-distribution sensors.}
    \vspace{2mm}
    \label{table:cross_sensor}

    \resizebox{\textwidth}{!}{
    \begin{tabular}{lcccccccc}
        \toprule
        & \multicolumn{2}{c}{\cellcolor{lightred}\textbf{Custom-built A (Seen)}}
        & \multicolumn{2}{c}{\cellcolor{lightred2}\textbf{Custom-built B (Unseen)}}
        & \multicolumn{2}{c}{\cellcolor{lightblue}\textbf{GelSight (Seen)}}
        & \multicolumn{2}{c}{\cellcolor{lightblue2}\textbf{GelSight (Unseen)}} \\
        \cmidrule(lr){2-3}
        \cmidrule(lr){4-5}
        \cmidrule(lr){6-7}
        \cmidrule(lr){8-9}
        \textbf{Task} 
        & \cellcolor{lightred}Ours & \cellcolor{lightred}Force Pred.
        & \cellcolor{lightred2}Ours & \cellcolor{lightred2}Force Pred.
        & \cellcolor{lightblue}Ours & \cellcolor{lightblue}Force Pred.
        & \cellcolor{lightblue2}Ours & \cellcolor{lightblue2}Force Pred. \\
        \midrule
        
        Whiteboard Erasing
        & \cellcolor{lightred}\textbf{60.0} & \cellcolor{lightred}\textbf{60.0}
        & \cellcolor{lightred2}\textbf{53.3} & \cellcolor{lightred2}13.3
        & \cellcolor{lightblue}\textbf{46.7} & \cellcolor{lightblue}40.0
        & \cellcolor{lightblue2}\textbf{33.3} & \cellcolor{lightblue2}6.67 \\
        
        Heavy Part Lifting
        & \cellcolor{lightred}\textbf{80.0} & \cellcolor{lightred}66.7
        & \cellcolor{lightred2}\textbf{67.7} & \cellcolor{lightred2}40.0
        & \cellcolor{lightblue}\textbf{73.3} & \cellcolor{lightblue}60.0
        & \cellcolor{lightblue2}\textbf{66.7} & \cellcolor{lightblue2}33.3 \\

        Tongs Food Pick
        & \cellcolor{lightred}\textbf{66.7} & \cellcolor{lightred}53.3
        & \cellcolor{lightred2}\textbf{46.7} & \cellcolor{lightred2}0.0
        & \cellcolor{lightblue}\textbf{53.3} & \cellcolor{lightblue}46.7
        & \cellcolor{lightblue2}\textbf{40.0} & \cellcolor{lightblue2}0.0 \\

        Chocolate Grasping
        & \cellcolor{lightred}\textbf{93.3} & \cellcolor{lightred}86.7
        & \cellcolor{lightred2}\textbf{73.3} & \cellcolor{lightred2}66.7
        & \cellcolor{lightblue}\textbf{86.7} & \cellcolor{lightblue}\textbf{86.7}
        & \cellcolor{lightblue2}\textbf{73.3} & \cellcolor{lightblue2}53.3 \\
        
        \bottomrule
    \end{tabular}}
\end{table*}

\subsection{Cross-sensor Robustness}\label{sec:exp:cross_sensor}

A fundamental design question in tactile perception is how to utilize the extracted information better. While our TaF-Adapter aligns tactile representations with force in a latent space, a commonly explored alternative is to explicitly regress force values from tactile images~\cite{chen2025transforce,fang2025force} and feed these predictions into the policy. To validate our design choice, we conducted a comparative study between two pipelines: 1) \textbf{TaF-Adapter (Ours):} The VLA policy receives high-dimensional latent tokens that are contrastively aligned with force dynamics; 2) \textbf{Force Prediction:} We train a supervised encoder-decoder on the TaF-dataset to regress 6-axis force/torque and matrix force maps directly from tactile images. These predicted values are then fed as input to the same VLA backbone. 
We evaluate both methods on four diverse tasks (\textit{Chocolate Grasping}, \textit{Heavy Part Lifting}, \textit{Tongs Food Pick}, and \textit{Whiteboard Erasing}), assessing both in-distribution performance and cross-sensor robustness. The results are summarized in \cref{table:cross_sensor}.

\paragraph{Performance on Seen Sensors}
We first evaluate performance on sensors encountered during pre-training (\textsc{Custom-built A} and \textsc{GelSight}, denoted as ``Seen'' sensors). As shown in \autoref{table:cross_sensor}, the TaF-Adapter consistently outperforms the explicit prediction baseline.
Specifically, on the \textsc{Custom-built A} sensor, the TaF-Adapter achieves an average success rate of 75.0\%, compared to 66.7\% for the force-prediction baseline (an absolute gain of $+8.3\%$). Similarly, on the \textsc{GelSight} sensor, our method achieves 65.0\% versus 58.3\% ($+6.7\%$).

\textbf{Analysis:} These results indicate that TaF-adapter generates representations that are more informative for downstream manipulation than the force-prediction strategy. We attribute this robustness to the adapter’s ability to preserve richer visual–tactile structure relevant to contact-rich manipulation, rather than collapsing the signal into low-dimensional force estimates that may discard task-critical information. In addition, task-specific fine-tuning enables the encoder to capture manipulation-relevant tactile cues, further improving its downstream effectiveness.

\paragraph{Cross-Sensor Generalization (Unseen Sensors)}

\begin{figure}[t!]
    \centering
    \includegraphics[width=\linewidth]{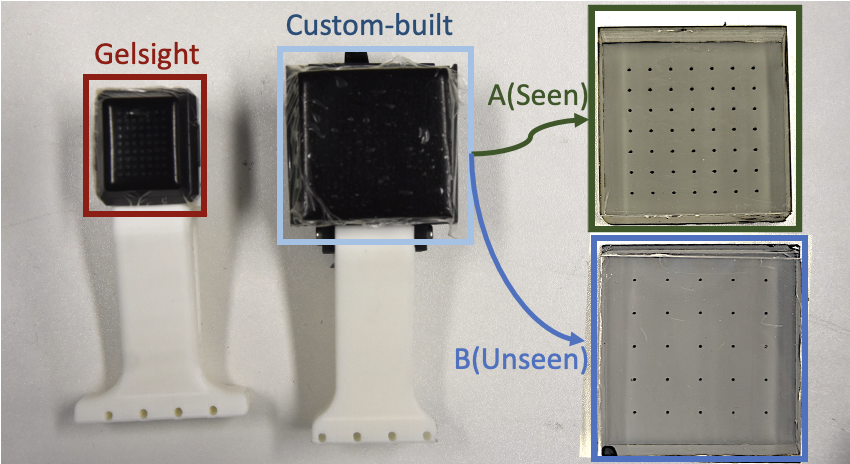}
    \caption{Three \ac{vbts} configurations.}
    \label{fig:sensors}
\end{figure}

We also highlight that our TaF-Adapter exhibits strong zero-shot generalization to sensors not encountered during pre-training, as shown in \autoref{table:cross_sensor}. When directly evaluated on the \textsc{Custom-built B (Unseen)} sensor, which did not appear in the TaF-Adapter’s pre-training phase (denoted as ``unseen''; see \cref{fig:sensors}). On \textsc{Custom-built B}, the TaF-Adapter maintains a robust mean success rate of 60.3\%. In sharp contrast, the force-prediction baseline suffers a catastrophic performance drop, falling to 30.0\%. A similar trend is observed on the unseen \textsc{GelSight} unit, where our method achieves 53.3\% compared to just 23.3\% for the baseline.

\textbf{Analysis:} Explicit force regression models are insensitive to minor domain shifts (\eg, lighting changes, gel wear, or marker displacement), as they rely on precise pixel-level mappings to output force values. When these mappings drift, the predicted force becomes noisy or erroneous, misleading the policy. The TaF-Adapter, however, aligns representations in a semantic latent space. This contrastive alignment encourages the model to learn force-relevant features that are invariant to low-level sensor noise, resulting in significantly superior cross-sensor transferability.

\subsection{Pretrained TaF-Adapter Improves Data Efficiency}\label{sec:exp:train_data}
\begin{figure}[t!]
    \centering
    \includegraphics[width=\linewidth]{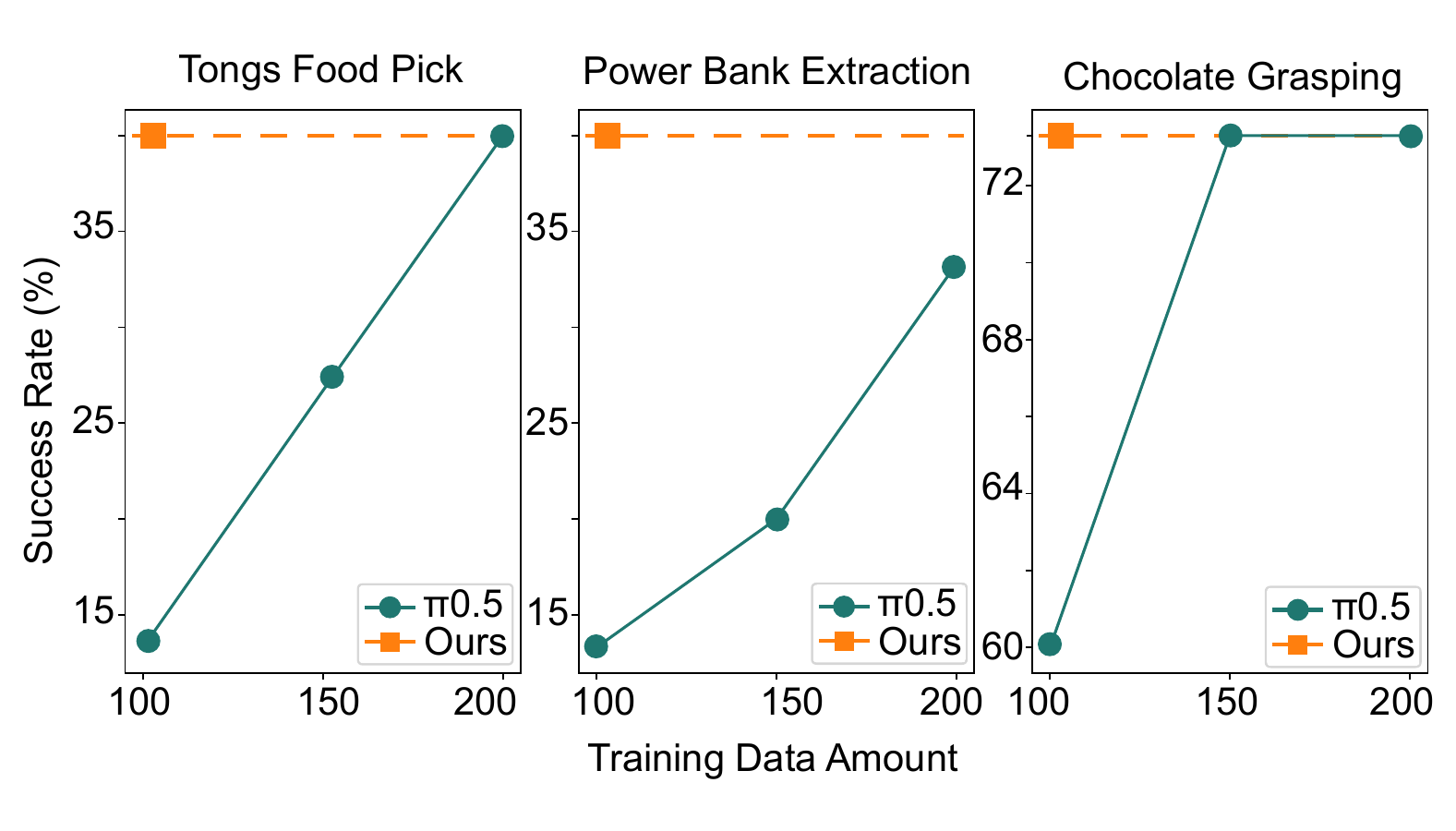}
    \caption{
    \textbf{Effect of training data amount on task performance.} 
    This figure compares the performance of TaF-VLA with that of the vision-only baseline $\pi_{0.5}$ under varying training data budgets.
    }
    \label{fig:train_efficiency}
\end{figure}

We further demonstrate that initializing the VLA backbone with our pre-trained TaF-Adapter yields substantial gains in \textit{data efficiency}. To quantify this, we evaluated \modelname against the vision-only baseline ($\pi_{0.5}$) across three representative contact-rich tasks (\textit{Tongs Food Pick}, \textit{Power Bank Extraction}, and \textit{Chocolate Grasping}), training on data subsets of 100, 150, and 200 demonstrations. The quantitative trends are visualized in \cref{fig:train_efficiency}.

As expected, the vision-only $\pi_{0.5}$ relies heavily on dataset scale: its success rate improves incrementally from 13.3\%~$\rightarrow$~40.0\% on \textit{Tongs Food Pick}, 13.3\%~$\rightarrow$~33.3\% on \textit{Power Bank Extraction}, and 60.0\%~$\rightarrow$~73.3\% on \textit{Chocolate Grasping} as the dataset increases from 100 to 200 episodes.
Crucially, \modelname achieves equivalent or superior performance using only 100 demonstrations---effectively matching the baseline's best performance while using half the data. This confirms that the force-aligned tactile codebook provides structured physical priors, significantly reducing the number of demonstrations required to learn complex, force-sensitive manipulation skills.

\subsection{Ablation Studies}\label{sec:exp:abalation}
\modelname relies on specific architectural choices to ensure robust alignment. To quantify the contribution of each component, we conducted ablation studies across three representative tasks: \textit{Whiteboard Erasing}, \textit{Tweezer Weight Pick}, and \textit{Jelly Slicing}. The results are summarized in \cref{tab:ablation_study}.

\textbf{Necessity of Temporal Context:}
We found that incorporating historical observations is essential for force-aware manipulation. Reducing the receding horizon window from $N=5$ to a single frame ($N=1$) leads to substantial performance degradation (refer to ``No History'' in \cref{tab:ablation_study}). This sensitivity arises because physical interaction is inherently history-dependent: a static grasp and a grasp undergoing incipient slip often yield identical instantaneous deformation images. Disambiguating these states requires temporal context to estimate the rate of change in the deformation field.

\textbf{Impact of Codebook Capacity:}
We find that the capacity of the codebook plays a critical role in overall model performance (refer to ``Small Codebook'' in  \cref{tab:ablation_study}). Reducing the codebook size leads to a decline in success rates. This degradation occurs because an undersized codebook forces different tactile–visual features to collapse into the same codebook entries, causing a loss of representational granularity. Consequently, the encoder fails to preserve subtle cues that are necessary to distinguish fine-grained interaction states.

\textbf{Benefits of Decoupled Modalities:}
Using two separate codebooks is also essential for achieving high success rates. To validate this, we evaluated a variant in which both the pressure maps and the global 6-DoF force vectors are projected into a single shared codebook (refer to ``Shared Codebook'' in  \cref{tab:ablation_study}). This configuration performs poorly across all three benchmark tasks.
This degradation underscores that global F/T vectors (low-dimensional, high-magnitude signals) and pressure maps (high-dimensional, spatial signals) exhibit fundamentally different latent distributions. Forcing them into a unified latent space causes modality-specific information to be aliased or lost. In contrast, our decoupled design preserves both the spatial geometry captured by pressure maps and the global dynamics encoded in force–torque vectors.

\textbf{Quality of Tactile Data:}
Model performance also depends heavily on the quality of tactile data collected by our hardware platform (\cref{sec:dataset:device}). When the force signals are perturbed, performance decreases substantially. To illustrate this effect, we trained a variant in which the ground-truth force labels were artificially corrupted with Gaussian noise. As shown by the ``w/ Noise'' configuration in \cref{tab:ablation_study}, this variant exhibits degraded performance. We attribute this drop to the difficulty of learning a meaningful alignment when the force modality is anchored to noisy or inconsistent labels. These results highlight the necessity of a stable and low-disturbance data acquisition pipeline, such as the one enabled by our device.

\begin{table}[t]
\centering
\caption{Ablation study results. The full pipeline achieves the highest overall performance, whereas each ablated variant exhibits a measurable reduction. }
\label{tab:ablation_study}
\renewcommand{\arraystretch}{1.1}
\resizebox{\columnwidth}{!}{%
\begin{tabular}{lccccc}
\toprule
\textbf{Task} 
& \textbf{\shortstack{No\\History}} 
& \textbf{\shortstack{Small\\Codebook}} 
& \textbf{\shortstack{Shared\\Codebook}} 
& \textbf{\shortstack{w/\\Noise}} 
& \textbf{\shortstack{Ours\\(Full)}}  \\
\midrule
Whiteboard Erase  & 26.7 & 46.7 & 13.3 & 13.3 & \textbf{60.0} \\
Weight Pick & 20.0 & 20.0 & 13.3 & 0.0  & \textbf{33.3} \\
Jelly Slice       & 26.7 & 20.0 & 6.7  & 6.7  & \textbf{53.3} \\
\bottomrule
\end{tabular}%
}
\end{table}

\section{Discussion and Limitations}\label{sec:discussion}
\subsection{Key Findings}
The experimental evaluation yields three critical insights regarding the integration of tactile perception into generalist policies:

\textbf{When Manipulation Requires Force Knowledge:} Our results show that the necessity of force sensing is highly task-dependent. For tasks involving lightweight objects where only gentle or bounded forces are required (\eg, \textit{Chocolate Grasping}), the vision-only baseline (\eg, $\pi_{0.5}$) performs comparably to tactile-based approaches.
However, for \textit{force-critical} tasks that require active modulation of contact forces (\eg, \textit{Heavy Part Lifting}, \textit{Power Bank Extraction}), relying solely on vision leads to significantly degraded performance. For example, in the \textit{Tweezer Weight Pick} task, the vision-only baseline fails completely regardless of how the language instructions are configured, as RGB observations do not contain information sufficient to infer stiffness, friction, or shear forces that are essential for successful manipulation.
In contrast, the TaF-Adapter overcomes this limitation by providing an explicit, force-aware representation that enables the policy to regulate physical interactions effectively.

\textbf{Discrete Representations Enable Generalization:}
Our evaluation shows that representing tactile features in a discrete space is critical for achieving strong generalization performance. The discrete codebook representation consistently outperforms continuous latent embeddings, despite the widespread use of continuous spaces in previous tactile learning research~\cite{yu2024octopi}. We found that discretization suppresses high-frequency noise and hardware-specific artifacts while retaining the essential physical cues needed for downstream reasoning. This is further supported by our zero-shot generalization results on sensors that are visually similar but structurally different, demonstrating that discrete quantization is a key mechanism for hardware-agnostic tactile perception.

\subsection{Limitations and Future Work}

\begin{figure*}[t!]
    \centering
    \includegraphics[width=\linewidth]{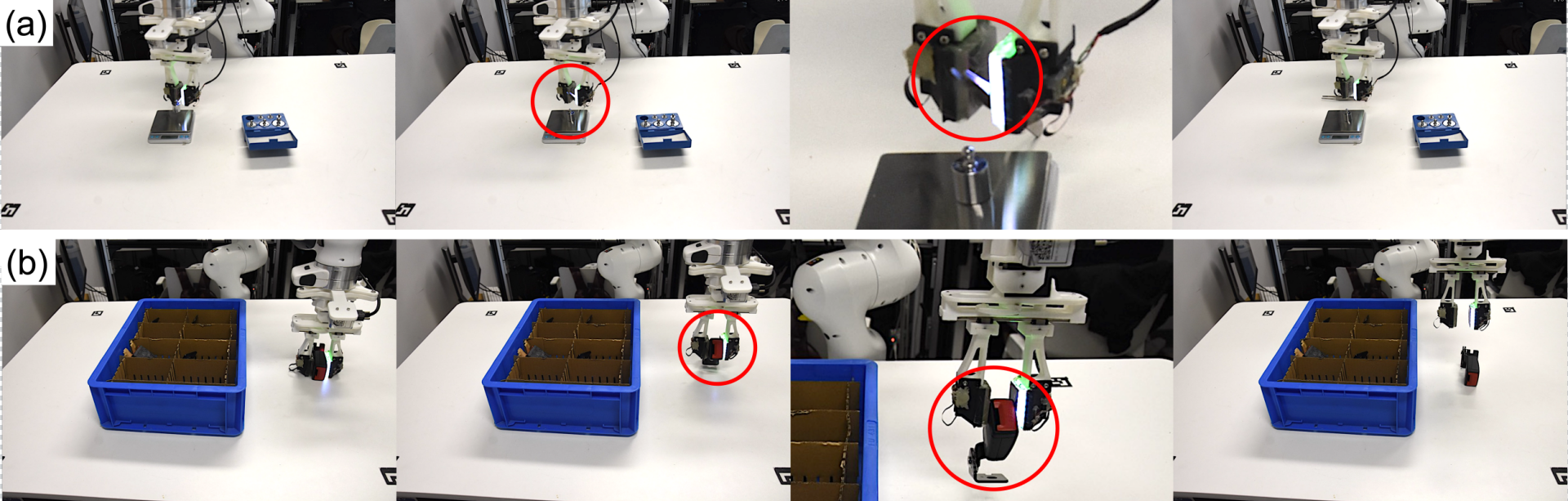}
    \caption{
    \textbf{Failure cases arising from insufficient force perception.} Examples highlight critical failures caused by inadequate grounding of physical forces:
    (a) The policy fails to maintain adequate clamping force, causing the tweezers to slip and drop the weight during transport.
    (b) When handling heavy objects, the system fails to exert sufficient force to counteract gravity, leading to the object dropping.}
    \label{fig:failure}
\end{figure*}

\textbf{Failure Modes:}
Failures were observed during our experiments. Most failures for \modelname occurred during the pre-contact phase. A predominant failure mode involves errors in manipulator control during free motion, where the manipulator fails to accurately reach the desired position. This is particularly critical in tasks requiring high precision, such as \textit{Tweezer Weight Pick}, where the end-effector stops millimeters short of the target surface. This “air gap” prevents the establishment of initial contact necessary to trigger the tactile–force feedback loop. Additionally, despite the involvement of the TaF-adapter, some failures still arise from insufficient force applied during the interaction process. Examples of these failure cases are illustrated in \cref{fig:failure}.

\textbf{Inference Frequency:}
Our approach leverages receding-horizon action output. While predicting a long horizon of future actions ensures trajectory smoothness, it inherently limits the frequency of closed-loop corrections. As a result, the policy exhibits latency and is prone to failure during high-frequency dynamic events, such as handling force spikes in \textit{Jelly Slicing}, where physical dynamics evolve faster than the reactive replanning cycle. Moreover, force interactions are inherently fast, highlighting the need for architectures capable of faster inference to meet the demands of high-frequency force control.

\textbf{Generalization Scope:}
Our approach demonstrates generalizability across sensors. However, the cross-sensor evaluation is limited to sensors that are relatively similar. The sensors we tested operate on the same visuotactile sensing mechanism and incorporate visual markers on their elastomer. It remains inconclusive how well the approach generalizes to sensors that are highly different, such as those based on capacitive, magnetic, or piezoresistive arrays, which rely on non-visual encoding mechanisms.

\textbf{Data Acquisition Hardware:}
The current TaF-Dataset is collected using custom TaF-device that are primarily fabricated through 3D printing. Due to the inherent precision constraints of consumer-grade additive manufacturing, the captured force signals and corresponding ground-truth measurements inevitably contain small structural and geometric deviations. These fabrication-induced inconsistencies may introduce minor biases into the learned tactile-force representations. Future iterations of the hardware could incorporate redesigned transmission mechanisms (\eg, direct drive) and high-precision machining (\eg, CNC milling or lathe finishing) to improve sensor fidelity and yield more accurate ground-truth force measurements.

\textbf{Alternative Architecture Designs:}
In our current design, the TaF-Adapter is integrated as input to the action expert. This choice facilitates easier fine-tuning and faster inference. However, it may not be optimal, as humans use tactile information not only to adjust their grip but also to update their understanding of the environment (\eg, realizing a surface is slippery and modifying their plan entirely). A model with deeper tactile integration could support far richer semantic reasoning. Therefore, exploring a fully unified Vision-Tactile-Language-Action backbone is meaningful. Nevertheless, we anticipate that this approach would require scaling up data collection efforts well beyond current limits and remains unexplored in the present work.

\section{Conclusion}\label{sec:conclusion}

In this paper, we present \modelname, a framework designed to address the fundamental ``force-blindness'' of current generalist robot policies. Our core insight is that directly using tactile images from \ac{vbts} may omit the underlying force priors they convey. To capture this force-semantic information, we propose the TaF-Adapter, a neural encoder trained to align \ac{vbts} latent codes with those from a separate force and pressure sensor array via contrastive learning.

To support this training, we developed an automated, scalable data acquisition device to collect tactile-force-aligned data, which are otherwise scarce. This resulted in the TaF-Dataset, providing over 10 million synchronized frames of \ac{vbts} images, 6D force/torque measurements, and matrix pressure maps---serving as a valuable resource for the research community.

We then demonstrated the integration of the TaF-Adapter into the \modelname VLA model and conducted comprehensive evaluations. \modelname achieves an average success rate of 64.8\%, roughly doubling the performance of state-of-the-art vision-only baselines and outperforming existing vision–tactile methods by 19–22 percentage points. Furthermore, we show that this force-aligned representation enables new capabilities in language-conditioned manipulation, allowing the robot to execute fine-grained force-modulation commands that were previously intractable. These results suggest that \modelname advances the acquisition of dexterous, generalist tactile-based skills and highlights the importance of explicitly grounding tactile perception with physical forces.

\begin{figure*}[th!]
    \centering
    \includegraphics[width=\linewidth]{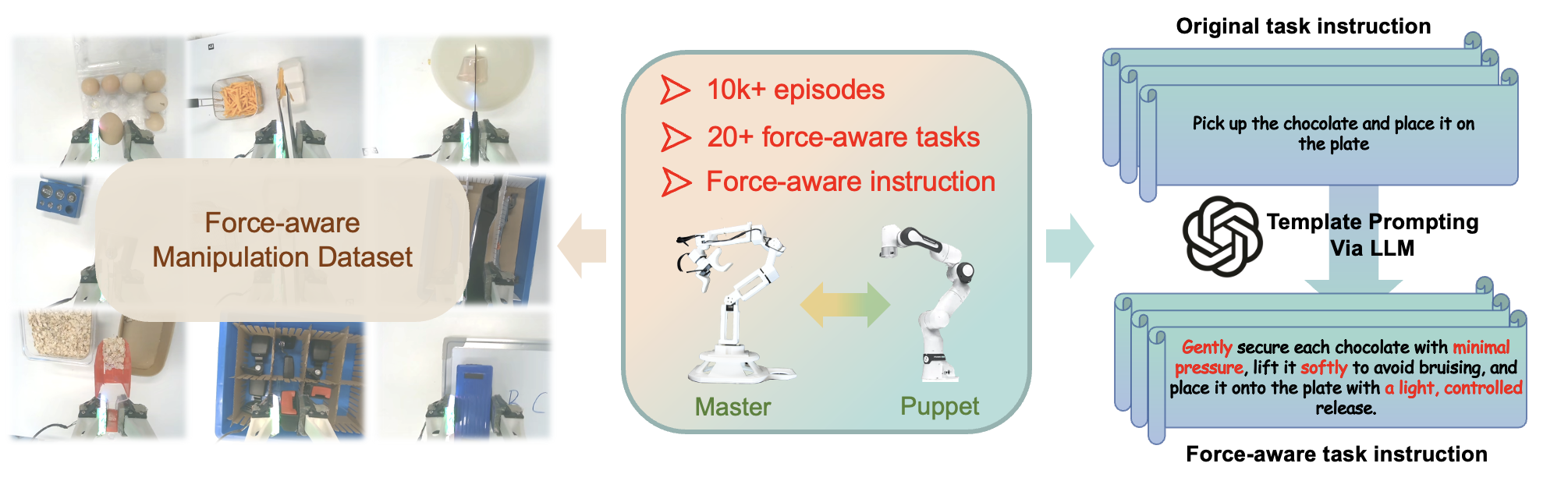}
    \caption{\textbf{Overview of the force-aware manipulation dataset construction pipeline.} We utilize a high-fidelity master-puppet teleoperation system (middle figure) to collect over 10,000 real-world manipulation episodes across 20+ diverse tasks, ranging from fragile object handling to tool use (left figure). To bridge the semantic gap between high-level goals and low-level control, we employ an LLM-based prompting framework (right figure) to generate force-aware task instructions. This process empowers force-aware directives, providing the hierarchical supervision necessary for force-aware downstream policy learning.}
    \label{fig:dataset}
    \vspace{-8pt}
\end{figure*}

\appendix

\subsection{Force-aware Manipulation Dataset Construction}\label{appendix:manipdataset}
To facilitate the learning of force-aware manipulation policies, we curate the force-aware manipulation dataset. As illustrated in \cref{fig:dataset}, the dataset included real-world demonstrations featuring high-fidelity visual and tactile modalities.

\subsubsection{Teleoperation Interface}
An overview of the data collection pipeline is presented in \cref{fig:device} (middle figure). We developed a teleoperation interface to control the arm motion. The system leverages a full-scale exoskeleton controller that is mechanically isomorphic to the Franka Research 3 (FR3) robot arm. Equipped with high-resolution joint encoders, the exoskeleton streams joint-space commands directly to the robot with low latency. 

\subsubsection{Selection of Force-aware Manipulation Tasks}
The dataset spans a broad range of physical interactions with objects of different stiffness, weight, and geometry, requiring the robot to modulate grip force according to object compliance and mass. Tasks include fragile items that demand minimal force to avoid damage and heavy, low-friction objects that require higher force to prevent slippage. Across these diverse conditions, the model must infer the appropriate force level from physical feedback to maintain a stable grasp.

In addition to direct grasping, the dataset includes tool-use scenarios such as using tweezers to lift weights. These tasks introduce extended kinematic chains in which the robot acts indirectly on the environment. The tool functions as a lever, amplifying rotational moments at the fingertips, so the policy must actively stabilize the tool during transport and reason about the dynamics of the coupled tool–hand system rather than a single contact point.

The dataset also contains tasks defined by dynamic surface interaction, where the objective is continuous regulation of contact rather than object retention. In the whiteboard-cleaning task, the robot must guide an eraser across the surface while maintaining steady normal force and compensating for shear to prevent skipping. In the precision-cutting task (slicing jelly on a balloon), the robot must detect the sharp change in resistance when the blade transitions from jelly to balloon and terminate the motion immediately. These scenarios evaluate the policy’s ability to operate as a responsive feedback controller, adapting its trajectory in real time as physical conditions change.

\subsubsection{Force-aware Instruction Generation}
A common limitation of existing manipulation datasets is the lack of language descriptions of tactile data. To address this, the force-aware manipulation dataset augments every demonstration with force-grounded language annotations.
We employ an automated annotation pipeline driven by a pretrained \ac{llm} via a physics-informed Chain-of-Thought (CoT) prompting strategy. The prompt template used for generating the TaF-Dataset is showcased in \cref{fig:prompt}.

\begin{figure}[t!]
    \centering 
    \framedtext{
    \textrm{\small 
    \vspace{-1mm} \\
    \textbf{System message}: You are a domain expert in robotic manipulation and physical interaction. Your task is to translate a high-level manipulation instruction into a concise, executable command that specifies how the action should be performed. Your output must use only abstract force descriptors (\eg, ``gently'', ``apply light pressure'', ``use minimal force'') and must never include numerical force values. Keep the final instruction brief, actionable, and focused on force-aware execution.\\
    \textbf{Original task instruction:} A robot arm is tasked to \texttt{Task Instruction}. \\
    \textbf{Physical Analysis:} Provide a concise assessment of the object’s key physical properties relevant to manipulation (\eg, fragility, stiffness, deformability, compliance, surface friction, mass distribution). Focus only on attributes that directly influence how the object should be handled. Keep this section brief and factual.\\
    \textbf{Strategy Derivation:}
    Provide a concise, high-level manipulation strategy that specifies how the object should be handled using only qualitative force descriptors. Do not include numerical forces, detailed step-by-step procedures, or multi-stage reasoning. Keep this strategy brief and focused on the underlying force principles.\\
    \textbf{Force-aware task instruction:}
    Rewrite the original instruction as a concise, execution-ready, and force-aware command.
    Use only abstract, qualitative force terms (\eg, ``gently secure'', ``apply minimal pressure'', ``lift softly''), and avoid enumerated steps, procedural detail, or any numerical force specification.\\
    \textbf{Output Format:} Please generate output following the JSON format below:
    \{
        "physical analysis": "...",
        "strategy derivation": "...",
        "force-aware instruction": "...",
    \}
    }
    }
    \vspace{-2mm}
    \caption{Prompt templates used for annotating the force-aware manipulation dataset.} 
    \label{fig:prompt}
    \vspace{-6pt}
\end{figure}

\subsection{Description of Task Suite}\label{appendix:task_analysis}
This section details the contact-rich manipulation tasks used in our evaluation.

\textbf{(a) Tube Insertion:}
Insert a test tube into a tight slot. As the tube is relatively slippery, the primary challenge is precise pose alignment. Force regulation is also required to prevent jamming caused by lateral contact forces. \textit{Success:} Full insertion without jamming or forceful collision.

\textbf{(b) Tweezer Weight Pick:}
A tool-use task involving tweezers that introduces an extended kinematic chain. The policy must manage delicate force transmission and counter-rotational moments at the tool tips. \textit{Success:} Transport and place at least two weights without dropping.

\textbf{(c) Power Bank Extraction:}
Extract a unit from a friction-fit slot. The task requires a stable grasp and precise directional pulling to overcome friction without binding or slip. \textit{Success:} Complete extraction and placement without slippage or collision.

\textbf{(d) Whiteboard Erasing:}
Guide an eraser along a specified trajectory while regulating normal and shear forces. Insufficient pressure causes intermittent contact; excessive force increases friction and may impede motion. \textit{Success:} Successfully remove all marks along the trajectory.

\textbf{(e) Jelly Slicing:}
Slice soft jelly placed atop a balloon. The robot must detect the stiffness transition via torque feedback and halt immediately upon reaching the balloon surface. \textit{Success:} Fully slice the jelly without rupturing the balloon.

\textbf{(f) Heavy Part Lifting:}
Lift and transport a heavy automotive component. The task requires a sustained high grip force to counter gravity and prevent incipient slip. \textit{Success:} Transport to the target location without slippage or loss of contact.

\textbf{(g) Tongs Food Pick:}
Manipulate irregular food items using compliant tongs. The policy must account for tool compliance and variable surface friction to maintain stability. \textit{Success:} Securely grip, transport, and place the item without dropping.

\textbf{(h) Chocolate Grasping:}
Grasp and transport a brittle chocolate object. The robot must balance slip prevention against fracture risk. \textit{Success:} Transport without breakage or surface damage.

\textbf{(i) Crisp Grasping:}
Grasp a crisp, representing an extreme fragility case with near-zero tolerance for force overshoot. \textit{Success:} Transport the crisp intact without cracking or shattering.


\bibliographystyle{ieeetr}
\bibliography{reference}

\end{document}